\documentclass{article}
\usepackage{aikyam}						

\usepackage{booktabs}						
\usepackage{multirow}						
\usepackage{amsfonts}	
\usepackage{amsmath}
\usepackage{amsthm}
\usepackage{graphicx}						
\usepackage{duckuments}						
\usepackage{tcolorbox}
\newtheorem{theorem}{Theorem}
\newtheorem{corollary}{Corollary}
\usepackage{tabularx}

\usepackage[numbers,compress,sort]{natbib}	

\title{Diminishing Returns of Data in Sparse Autoencoders: Functional Behavior does not Guarantee Feature Reproducibility}

\newcommand{\xhdr}[1]{\vspace{0em}\noindent{{\bf #1.}}}

\newcommand{\hide}[1]{}

\usepackage{colortbl}

\definecolor{Gray}{gray}{0.9}
\definecolor{LightCyan}{rgb}{0.88,1,1}
\definecolor{darkred}{rgb}{0.8,0.1,0.1}
\definecolor{darkyellow}{rgb}{0.95, 0.68, 0.22}
\definecolor{darkgreen}{rgb}{0.1,0.8,0.1}
\newcolumntype{a}{>{\columncolor{Gray}}c}
\newcolumntype{b}{>{\columncolor{white}}c}

\author[Ghosh et al.]{
Sambit Ghosh\textsuperscript{1,2*} \quad
Madelyn Mathai\textsuperscript{3*} \quad
R. Venkatesh Babu\textsuperscript{2} \quad
Chirag Agarwal\textsuperscript{3} \\
\textsuperscript{1}IBM 
\textsuperscript{2}Indian Institute of Science 
\textsuperscript{3}University of Virginia \\
\texttt{sambitghosh@iisc.ac.in}
}

\begin{document}

\maketitle

\begin{abstract}
    
\looseness=-1 Sparse autoencoders (SAEs) are typically trained on hundreds of millions to
billions of activation tokens, yet little work has tested whether this scale is
necessary. We study how much activation data SAEs need by training four SAE
architectures on three language models with budgets ranging from 5M tokens to a
500M-token reference. We measure two notions of convergence: functional behavior
on standard evaluations and one-to-one reproducibility of learned features.
Functional data requirements are strongly metric-dependent: reconstruction often
approaches the full-data reference at substantially smaller budgets, whereas
feature reproducibility continues to improve with data. For Gemma-2-2B
BatchTopK, the fraction of strictly matched features rises from 2\% at 5M tokens
to 29\% at 50M. Data selection also matters: random sampling better preserves
the original activation distribution and generally favors reconstruction,
whereas diversity-based sampling frequently improves AutoInterp. A theoretical
analysis separates statistical consistency of subsampled training from the
stronger identifiability and optimization conditions required for one-to-one
feature recovery. Our results show that SAE training can often be made far more data-efficient when budgets are matched to the goal, and feature-level claims are backed by explicit reproducibility checks, paving the way toward cheaper and more reliable SAE research. The code can be found on \href{https://github.com/AikyamLab/data-sae}{here}.    
\end{abstract}

\section{Introduction}

\looseness=-1 Sparse autoencoders (SAEs) have become a widely used tool for mechanistic interpretability, aiming to decompose dense model activations into sparse, interpretable features~\cite{bricken2023monosemanticity, cunningham2023sparse, sharkey2025open}. A central motivation is \emph{polysemanticity}: individual neurons can respond to multiple unrelated concepts, making neuron-level interpretation difficult~\citep{olah2020zoom, elhage2022superposition}. SAEs address this by learning an encoder that maps activations to sparse latent features and a decoder that reconstructs original activations~\citep{olshausen1996emergence}. Recent work has scaled SAEs to frontier models such as GPT-4~\citep{gao2024scaling} and Claude 3 Sonnet~\citep{templeton2024scaling}, motivating training on increasingly \textbf{large collections of activations}.

\looseness=-1 This raises a basic question: \emph{how much activation data does an SAE actually need?} We study this \textit{unexplored} question in an online training setting by varying activation exposure and measuring two notions of convergence: \textit{functional behavior} asks how reconstruction and downstream SAE metrics change with activation budget, while \textit{feature convergence} asks when the learned features themselves become reproducible under one-to-one matching. 
To clarify the distinction, we develop a theoretical framework of when additional data will lead to feature recovery and show that reducing the amount of activation data alone is not sufficient to guarantee dictionary convergence. This distinction matters because a central motivation for SAEs is not merely good reconstruction but reliable feature discovery. Prior work has questioned whether independently trained SAEs recover the same latent structure~\citep{paulo2025same, gerasimov2026unstable, lan2024universality}, with \citet{paulo2025same} using Hungarian matching~\citep{kuhn1955hungarian} and encoder-decoder agreement to quantify feature correspondence across dictionaries.

Our work suggests that large activation budgets are not universally required for high-quality SAE behavior.
Our main contributions are: 1) We derive a theoretical characterization of feature recovery under data reduction, and separate statistical consistency from the additional requirements of feature identifiability and successful optimization; 2) across three language models, four SAE architectures, multiple configurations, and activation budgets we present a systematic study of online SAE data requirements; 3) we demonstrate a distinction between functional behavior and feature convergence, where functional SAE metrics exhibit distinct data-scaling behavior, while one-to-one feature reproducibility continues to improve substantially with additional activation exposure; and 4) we show that data selection changes SAE quality in metric-dependent ways: random sampling better preserves reconstruction and intervention-level metrics, while diversity-based sampling shifts training toward high-norm activations and frequently improves Autointerp~\cite{paulo2024automatically}.


\looseness=-1 To the best of our knowledge, our findings provide the \textbf{first systematic evidence} that SAE training can be \textbf{substantially more} data-efficient than current large-scale practice. For reconstruction and several functional evaluations, small activation budgets (5M-token) recover much of the behavior of a 500M-token SAE. In contrast, reproducible feature recovery becomes increasingly reliable with larger activation budgets, highlighting SAE's data requirements depend on the evaluation objective.

\section{Related Work}
\xhdr{Sparse autoencoders and evaluation} Since the standard SAE~\citep{bricken2023monosemanticity, cunningham2023sparse}, several architectural variants have been proposed, including JumpReLU~\citep{rajamanoharan2024jumping}, TopK and BatchTopK~\citep{gao2024scaling, bussmann2024batchtopk}, and Matryoshka SAEs~\citep{bussmann2025matryoshka}. Evaluation has likewise moved beyond reconstruction loss to feature absorption~\citep{chanin2024absorption}, sparse probing~\citep{gurnee2023finding, gao2024scaling}, unlearning~\citep{farrell2024applying}, and automated interpretability~\citep{bills2023language, paulo2024automatically}, which \citet{karvonen2025saebench} compiled into SAEBench, now the standard benchmark for SAE evaluation.

\looseness=-1\xhdr{Feature stability} A separate line of work asks whether the features SAEs recover are stable. \citet{paulo2025same} find that SAEs trained on the same data with different seeds learn substantially different features, casting doubt on whether any single SAE identifies a canonical set. Yet \citet{lan2024universality} show that the spaces these features span remain similar, even across language models. \citet{gerasimov2026unstable} reconcile these findings: Seed-unstable features carry little reconstruction or predictive signal and cluster in reproducible low-rank subspaces, suggesting basis ambiguity rather than noise. We complement this work by asking how feature reproducibility depends on the amount of activation data.

\looseness=-1\xhdr{Data selection} Outside interpretability, data selection is a well-studied route to efficiency: coreset methods seek a small subset that preserves the training signal of the full dataset~\citep{moser2025coreset}, and \citet{coleman2020selection} show that cheap proxy models can select such subsets an order of magnitude faster without hurting final accuracy. A common diversity-based strategy is farthest point sampling, which greedily adds the point farthest from the current selection~\citep{gonzalez1985clustering, eldar1997farthest} and has been applied to core-set selection for deep networks~\citep{sener2018active}. Closest to our setting, \citet{yang2025diversity} use SAE features to prune instruction-tuning data, but they select the language model's training data rather than the SAE's. We instead ask whether the choice of activations matters beyond their quantity.

\textit{Research gap:} Standard practice trains SAEs on hundreds of millions of tokens~\citep{karvonen2025saebench, lieberum2024gemma}. To our knowledge, no prior work systematically tests whether this scale is necessary, or whether a selected subset of activations can match SAEs trained on far larger activation budgets on established metrics.
\section{When Should Feature Matching Agree?} 
\label{sec:feature_alignment_theory}

Let $\widehat{\theta}_{1,N}$ denote an SAE trained on the full activation sample and $\widehat{\theta}_{q,N}$ denote an SAE trained on a subsample of value-independent random activations with fraction $q\in(0,1)$. Both runs use the same SAE training objective. Let $R(\theta)=\mathbb{E}_{X\sim P_X}[\ell_\theta(X)]$ denote the corresponding population objective, where $\ell_\theta(X)$ represents the training loss the SAE computes on a single activation X, using parameters $\theta$, and $\widehat R_{1,N}(\theta) = \frac{1}{N} \sum_{t=1}^{N} \ell_\theta(X_t)$ denote full sample empirical risk. We define the indices retained by value-independent random sampling using \(S_{q,N}\subseteq\{1,\ldots,N\}\) and the subsample empirical risk using $\widehat R_{q,N}(\theta) = \frac{1}{|S_{q,N}|} \sum_{t\in S_{q,N}} \ell_\theta(X_t).$ Finally, $E(\theta) = [e_1(\theta),\ldots,e_p(\theta)]$ and $D(\theta) = [d_1(\theta),\ldots,d_p(\theta)]$ denote the encoder- and decoder-side feature system, respectively. We assume that features are normalized before cosine-based comparison. Let $\mathcal S_p$ denote the set of $p\times p$ permutation matrices, so that for $P\in\mathcal S_p$, $\bar E(\theta')P$ is $\bar E(\theta')$ with its columns reordered. For two SAE parameter settings $\theta$ and $\theta'$, define the encoder and decoder one-to-one distances as
\begin{align}
d_E(\theta,\theta')
&=
\min_{P\in\mathcal S_p}
\left\|
\bar E(\theta)
-
\bar E(\theta')P
\right\|_F,
\\
d_D(\theta,\theta')
&=
\min_{P\in\mathcal S_p}
\left\|
\bar D(\theta)
-
\bar D(\theta')P
\right\|_F,
\end{align}
where $\|\cdot\|_F$ is the Frobenius norm, which is exactly zero when the two systems have the same feature directions.


\xhdr{Assumptions}
We require four conditions:
\textit{(i) Common population objective:} full- and reduced-data runs use the
same architecture, preprocessing, sparsity mechanism, loss weights, and
population weighting;
\textit{(ii) Statistical consistency:} both empirical objectives converge
uniformly to the same population risk;
\textit{(iii) Optimization consistency:} training approaches the empirical optimum as the number of observed activations increases; and
\textit{(iv) Feature identifiability:} all population minimizers recover the
same encoder and decoder feature systems up to a common latent permutation,
with positive separation from alternative feature systems.

\begin{theorem}[Asymptotic encoder and decoder feature alignment]
\label{thm:encoder_decoder_alignment}
Under the assumptions above, for every fixed $q>0$,
\begin{align}
d_E(
    \widehat\theta_{q,N},
    \widehat\theta_{1,N}
)
&\xrightarrow{p}0,
\\
d_D(
    \widehat\theta_{q,N},
    \widehat\theta_{1,N}
)
&\xrightarrow{p}0.
\end{align}
Equivalently,
\begin{align}
\min_{P_E\in\mathcal S_p}
\left\|
\bar E_{q,N}
-
\bar E_{1,N}P_E
\right\|_F
&\xrightarrow{p}0,
\\
\min_{P_D\in\mathcal S_p}
\left\|
\bar D_{q,N}
-
\bar D_{1,N}P_D
\right\|_F
&\xrightarrow{p}0.
\end{align}
\end{theorem}

\textit{Proof sketch.}Under assumptions (i)–(ii), the full- and reduced-data empirical risks converge uniformly to $R$. By (iii), both trained SAEs approach the minimizers of $R$, and by (iv) those minimizers share a single feature system up to permutation. The result follows by the triangle inequality. 
\begin{corollary}[Agreement of encoder and decoder matchings]
\label{cor:matching_agreement}
Let $P_E^{(N)}$ and $P_D^{(N)}$ denote the permutations that maximize total cosine similarity between the two SAEs' encoder and decoder directions, respectively:
\begin{align}
P_E^{(N)} &= \arg\max_{P\in\mathcal S_p}\ \sum_{j=1}^{p}\cos\!\big(e_{q,N,j},\, e_{1,N,P(j)}\big), \\
P_D^{(N)} &= \arg\max_{P\in\mathcal S_p}\ \sum_{j=1}^{p}\cos\!\big(d_{q,N,j},\, d_{1,N,P(j)}\big).
\end{align}
Suppose that at the population optimum the encoder and decoder directions of each feature share the same index, and that the optimal one-to-one matching is unique for both encoder and decoder cosine sums, exceeding every other permutation by a fixed margin $\delta>0$. Then, under the assumptions of Theorem~\ref{thm:encoder_decoder_alignment},
\[
\Pr\!\big(P_E^{(N)} = P_D^{(N)}\big) \longrightarrow 1
\qquad\text{as } N\to\infty .
\]
\end{corollary}

\begin{corollary}[Assignment agreement is weaker than strong feature recovery]
\label{cor:assignment_vs_similarity}
\looseness=-1 Encoder--decoder assignment agreement does not imply high encoder and decoder cosine similarity at a finite data budget. In particular,
\[
P_E^{(N)}(j)
=
P_D^{(N)}(j)
=
k
\]
only states that feature $k$ is independently selected as the optimal
one-to-one counterpart under both encoder and decoder matching. It does not
imply $\cos \left(e_{q,N,j}, e_{1,N,k} \right) > \tau$ or $\cos \left(d_{q,N,j}, d_{1,N,k} \right) > \tau$ for a fixed threshold $\tau$. Thus, substantial agreement between encoder- and decoder-based Hungarian assignments can coexist with a much smaller fraction of strongly aligned features.
\end{corollary}

\looseness=-1
\xhdr{Interpretation}
Theorem~\ref{thm:encoder_decoder_alignment} characterizes conditions under
which reducing activation data preserves the asymptotic feature system. It also
clarifies that, at finite budgets, assignment agreement is weaker than strong
directional recovery: encoder and decoder matchings may select the same reference
counterpart even when their cosine similarities remain moderate. This distinction
motivates our empirical comparison of counterpart agreement and strict
shared-feature recovery in Sec.~\ref{sec:feature_reproducibility}.

\section{Experimental Setup}
\label{experimental}

\looseness=-1 We study SAE data requirements along two axes: \emph{activation quantity} and \emph{activation selection}. Unless stated otherwise, we follow the corresponding SAEBench~\citep{karvonen2025saebench} training and evaluation configurations so that small-data runs are directly comparable to the large-data reference.

\xhdr{Models and SAE Configurations} We evaluate three large language models (LLMs) - Gemma-2-2B~\citep{gemma2024gemma2} (layer 12), Pythia-160M~\citep{biderman2023pythia} (layer 6), and Qwen3-8B ~\citep{yang2025qwen3}(layer 17), and four SAE architectures: BatchTopK~\citep{bussmann2024batchtopk}, Matryoshka BatchTopK~\citep{bussmann2025matryoshka}, JumpReLU~\citep{rajamanoharan2024jumping}, and ReLU\citep{bricken2023monosemanticity, cunningham2023sparse}, across multiple configurations. Within each comparison, we hold the LLM, activation site, dictionary width, sparsity setting, and optimization hyperparameters fixed, varying only the activation budget or selection strategy.

\looseness=-1\xhdr{Online SAE Training} SAEs are trained on FineWeb~\citep{penedo2024fineweb} in an \emph{online} activation-generation setting~\cite{marks2024dictionary_learning}. Text is streamed through a frozen LLM, target-layer activations are buffered, and SAE mini-batches are drawn from the selected activations before the buffer is refreshed with new text. Selected activations are used once, avoiding a fixed precomputed dataset. Across budgets, we fix the \(500\)M training configuration and optimization steps \cite{marks2024dictionary_learning}, varying only the retained activation fraction to isolate data scale from training exposure. 

\xhdr{Training-Data Budgets} The reference configuration is designed for \(500\)M activation tokens using ~\cite{marks2024dictionary_learning} framework. For reduced-data experiments, we retain a fraction $q\in\{0.01,0.05,0.10,1.0\}$ of the activations generated at each buffer refresh. Thus, each fresh activation pool contributes \(1\%\), \(5\%\), \(10\%\), or \(100\%\) of its activations to SAE training. The selected activations are consumed once before a new activation pool is generated from the online data stream. Therefore, \(q\) controls the amount of fresh activation data retained from each online buffer, rather than the size of a fixed replay dataset.

\looseness=-1\xhdr{Data sampling} We reduce the activation set using two sampling strategies: random sampling and farthest-point sampling (FPS), which explicitly promotes geometric diversity.





\looseness=-1\xhdr{Evaluation} We evaluate all SAEs using SAEBench~\citep{karvonen2025saebench} default configurations and metrics, including reconstruction, sparse probing~\citep{gurnee2023finding, gao2024scaling}, absorption~\citep{chanin2024absorption}, SCR~\citep{marks2024sparse}, TPP~\citep{karvonen2025saebench}, RAVEL~\citep{huang2024ravel}, automated interpretability~\citep{bills2023language, paulo2024automatically}, and unlearning~\citep{farrell2024applying}. We measure feature reproducibility against the \(500\)M reference using one-to-one Hungarian matching of encoder and decoder features~\citep{paulo2025same,kuhn1955hungarian}. Matching similarity and encoder-decoder assignment agreement distinguish \emph{feature convergence} from \emph{functional convergence}

\section{Functional Data Requirements Are Metric Dependent}
\label{sec:functional_saturation}

\begin{figure}[t]
    \centering
    \includegraphics[width=0.9\columnwidth]
    {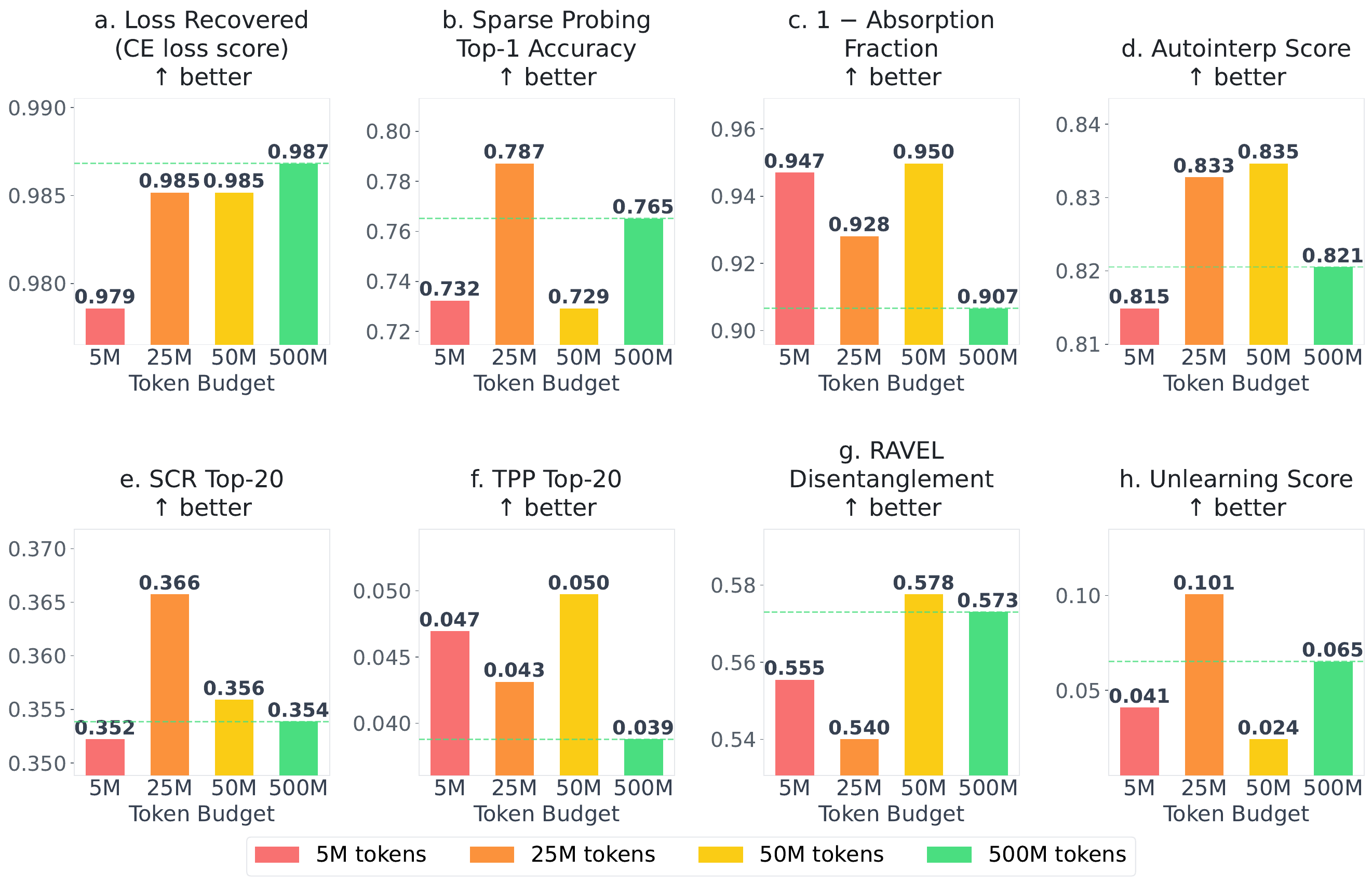}
    
    \caption{\looseness=-1\textbf{Functional data scaling for a BatchTopK SAE with \(16\)k latents.} SAEBench metrics are reported for SAEs trained with \{5M, 25M, 50M, 500M\} activation tokens. While smaller activation budgets approaches the 500M reference for reconstruction, probing, AutoInterp, and intervention-metrics (e, h) exhibit heterogeneous, non-monotonic dependence on activation exposure.}\label{fig:batch_top_k_16k_functional}
    \vspace{-0.43cm}
\end{figure}

\begin{figure}[t]
    \centering
    \includegraphics[width=0.9\columnwidth]
    {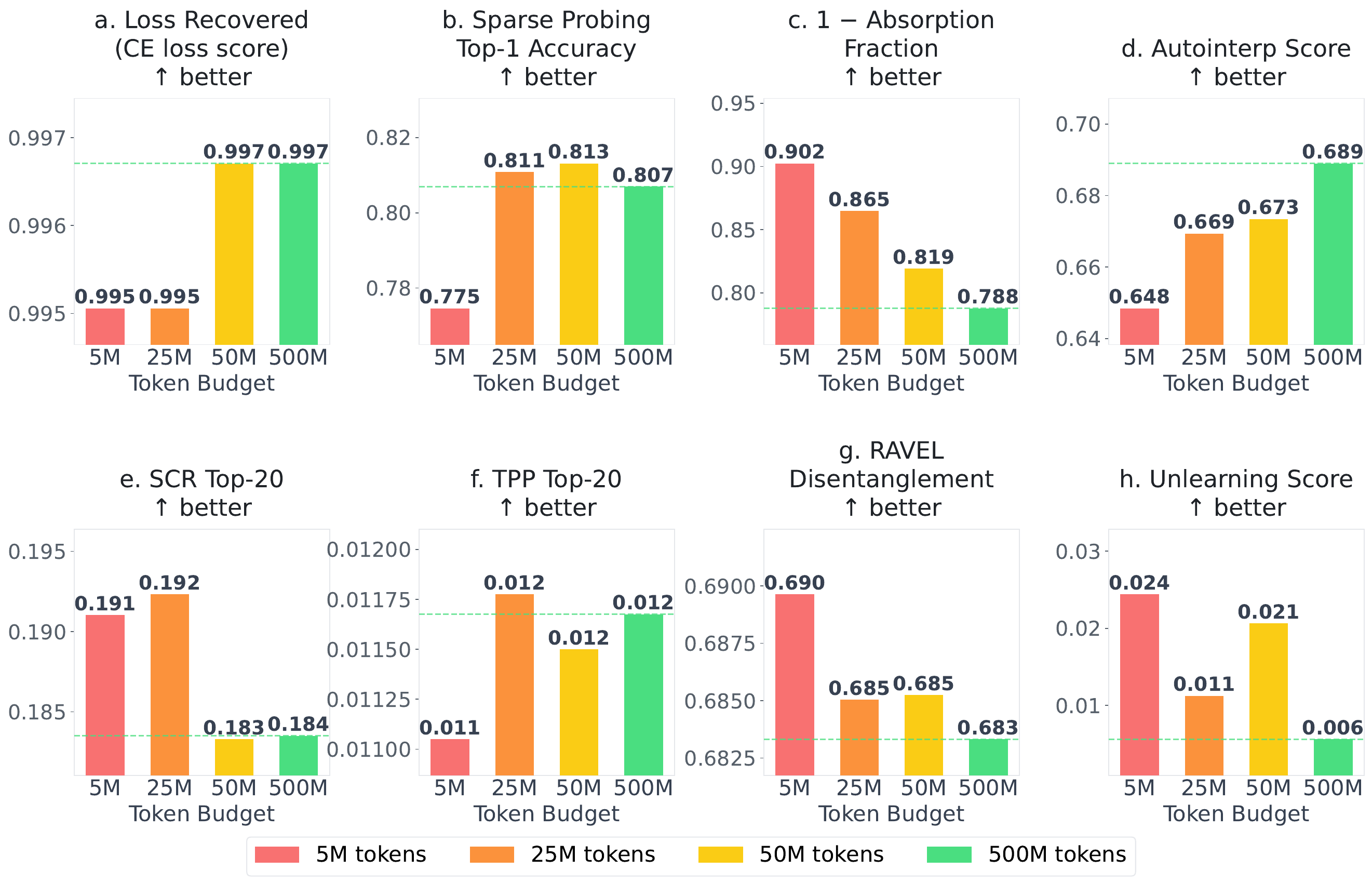}
    \caption{
    \looseness=-1\textbf{Functional data scaling for a standard ReLU SAE with \(65\)k latents.} SAEBench metrics are reported for SAEs trained with \{5M, 25M, 50M, 500M\} activation tokens. Reconstruction remains close to the \(500\)M reference at smaller activation budgets, whereas feature-level metrics show metric-dependent and frequently non-monotonic responses to increasing activation exposure.
    }
    \label{fig:standard_relu_65k_functional}
    \vspace{-0.4cm}
\end{figure}
Here, we show how much activation data is required to reproduce the functional behavior of a large-data SAE. 
We compare SAEs trained on randomly sampled activation budgets of \(5\)M, \(25\)M, and \(50\)M tokens against corresponding \(500\)M-token references. We focus on two representative configurations: a BatchTopK SAE with \(16\)k latents and a standard ReLU SAE with \(65\)k latents trained on layer~12 residual-stream activations of Gemma-2-2B. Figs.~\ref{fig:batch_top_k_16k_functional}-\ref{fig:standard_relu_65k_functional} report the corresponding SAEBench metrics across activation budgets. Results for additional SAE architectures, dictionary widths, and sparsities are provided in Appendix~\ref{appendix:functional_saturation}. 

\looseness=-1\xhdr{a. Reconstruction} The loss measures how well the SAE reconstruction preserves the model's next-token prediction behavior relative to the original model and a zero-ablation baseline. 

\textit{\underline{Observation:} The information required for reconstruction is learned early.} In Figs.~\ref{fig:batch_top_k_16k_functional}-\ref{fig:standard_relu_65k_functional}, we find that all reduced-data SAEs remain close to the \(500\)M reference. While the \(5\)M setting shows only a modest gap, \(25\)M and \(50\)M SAEs are generally very close to the reference,
highlighting that substantially less than \(500\)M activation tokens are sufficient to retain the LLM's behavior captured by SAE reconstruction.

\xhdr{b. \(k\)-Sparse Probing (SP)} It asks whether a pre-specified concept can be recovered from a small number of SAE latents. Following SAEBench, we report \(k=1\), so the metric specifically probes whether concept information is concentrated in an individual latent. 

\textit{\underline{Observation:} Concept localization is largely preserved at smaller data budgets.} Across sparsity settings and dictionary widths, reduced-data SAEs remain close to the \(500\)M reference. Sparse-probing accuracy is broadly consistent across \(5\)M-\(500\)M budgets, with smaller budgets sometimes outperforming the \(500\)M reference, showing that concept localization is well preserved with substantially less data.

\xhdr{c. Feature Absorption (FA)} It measures a complementary failure mode: whether a concept that should be represented by its main feature is instead partially captured by other latents. We report \(1-\) mean absorption; higher is better. 

\textit{\underline{Observation:} More data does not systematically reduce concept fragmentation.} FA is more sensitive to data budget than SP, yet reduced-data SAEs frequently outperform the \(500\)M reference. Together, SP and FA show that strong concept localization and feature alignment can be achieved with substantially smaller data budgets.

\looseness=-1\xhdr{d. Automated Interpretability} AutoInterp asks an LLM to infer a semantic description of a latent from its activating examples and evaluates whether that description predicts activation on held-out sequences, probing the semantic description of individual features rather than reconstruction quality. 

\textit{\underline{Observation:} Feature description is not monotonically determined by activation quantity.} AutoInterp varies moderately with activation budget, but its ordering depends on SAE configuration, including sparsity and dictionary width. Reduced-data SAEs outperform the \(500\)M reference in many regimes and underperform it in others. In several settings, the trend is even reversed, with smaller activation budgets yielding higher AutoInterp scores. Thus, increasing activation exposure \textbf{does not} consistently improve the semantic description of individual features.



\xhdr{e. Spurious Correlation Removal (SCR)} It evaluates whether ablating a small set of concept-related latents removes a spurious correlation while preserving the intended classification signal. 

\textit{\underline{Observation:} SCR shows moderate but non-monotonic sensitivity to data.} SCR varies more with activation budget than reconstruction, but increasing data does not consistently improve intervention specificity, and the performance of reduced-data SAEs depends on the width and the architecture.

\xhdr{f. Targeted Probe Perturbation (TPP)} It asks whether ablating latents associated with one class primarily affects that class while leaving others unchanged. 

\textit{\underline{Observation:} TPP is remarkably insensitive to activation budget} Across both dictionary sizes, TPP remains close to the \(500\)M reference despite the large difference in activation exposure. The ordering of data budgets is non-monotonic, but the absolute deviations are small, suggesting that this measure of class-specific feature separation requires relatively little additional data.

\xhdr{g. RAVEL} It evaluates whether an intervention successfully changes a targeted attribute while preserving unrelated attributes. 

\textit{\underline{Observation:} RAVEL is non-monotonic across activation budgets.} It varies with activation budget and architecture, indicating that effective intervention can be achieved without requiring consistently larger activation exposure.

\xhdr{h. Unlearning} It suppresses latents associated with a target domain and measures whether target knowledge can be removed while unrelated performance is preserved.

\textit{\underline{Observation:} Unlearning is the most configuration-sensitive evaluation} It varies substantially with activation budget, architecture, and sparsity, highlighting the importance of these factors for unlearning performance. Reduced-data SAEs span a wide performance range relative to the \(500\)M reference, highlighting that effective unlearning can emerge even with substantially smaller activation budgets. To strengthen all the observations above, we additonally add a paired statistical test in Table~\ref {tab:stats_gemma}.



In total, these results raise a natural next question: if all other hyperparameters are held constant, are the observed volatile differences primarily a consequence of using less data, or do they also depend on \emph{\textbf{which}} activations are randomly selected? To separate data quantity from data selection, we next compare random sampling with diversity-based activation selection at matched budgets.

\noindent{\textbf{Does Data Selection Matter?}} 
Here, we ask whether SAE data requirements are determined only by \emph{how much} data is used, or also by \emph{which} activations are selected. To separate these effects, we compare uniform random sampling (RS) with Farthest Point Sampling  (FPS)~\citep{gonzalez1985clustering, eldar1997farthest} at matched activation budgets. While RS retains activations independently of their values and therefore preserves the frequency structure of the activation stream in expectation, FPS selects activations to increase geometric coverage, preferentially retaining points that are far from those already selected. We follow the evaluation configuration of Figs.~\ref{fig:batch_top_k_16k_functional}-\ref{fig:standard_relu_65k_functional} and compare both selection strategies.
Results for additional SAE architectures are reported in Appendix~\ref{appendix:functional_saturation}.

\begin{figure}[t]
    \centering
    
    \includegraphics[width=0.95\columnwidth]{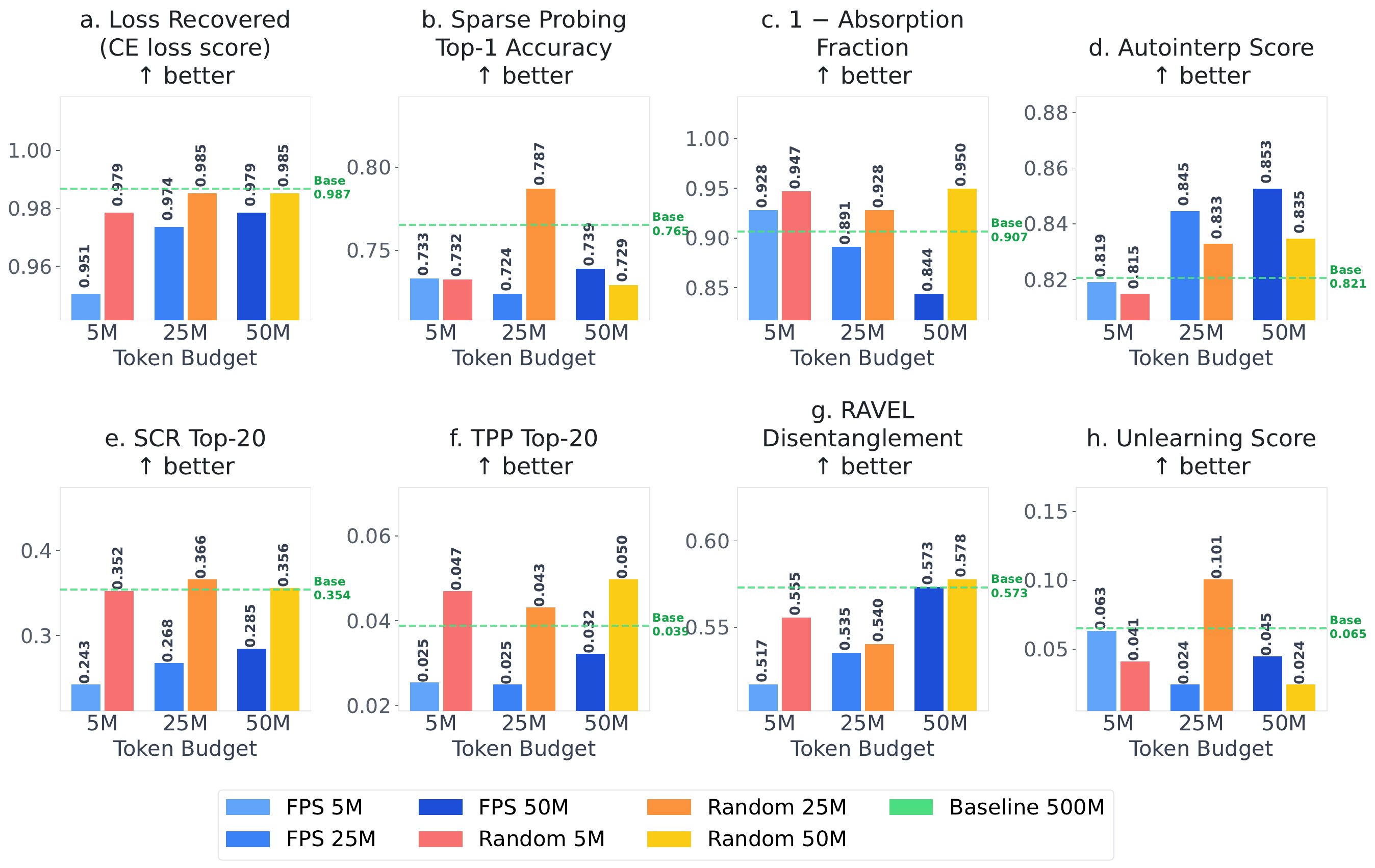}
    \caption{\looseness=-1 \textbf{Random sampling (RS) vs. FPS for a BatchTopK SAE with \(16\)k latents.} At each activation budget, paired bars compare FPS and RS, while the green dashed line denotes the corresponding \(500\)M reference. Selection strategy induces metric-dependent tradeoffs: RS better preserves reconstruction and several intervention metrics, whereas FPS often improves AutoInterp.}
    \label{fig:selection_16k}
    \vspace{-0.6cm}
\end{figure}

Fig.~\ref{fig:selection_16k} shows that the choice of activation subset matters even at a fixed budget: RS and FPS do not produce uniformly better or worse SAEs, but instead favor different properties of the learned representation.

\xhdr{\textit{i) Random sampling preserves reconstruction}} Across dictionary sizes and budgets, loss recovered under RS stays closer to the \(500\)M reference than under FPS, most visibly at small budgets. Geometric coverage alone thus preserves the model's predictive behavior less faithfully than representative sampling.

\xhdr{\textit{ii) Concept- and intervention-level metrics favor RS}} SCR, TPP, and RAVEL also favor RS across budgets, in some cases exceeding the \(500\)M reference, while SP and FA show no consistent winner.

\xhdr{\textit{iii) FPS yields higher AutoInterp}} FPS outperforms RS on AutoInterp at every budget in both configurations, and in 79\% of matched comparisons across all architectures, models, and budgets. FPS shifts training toward geometrically distinct, high-norm activations and is associated with higher AutoInterp but lower reconstruction scores. FPS also outperforms RS on unlearning at some budgets (5M and 50M in Fig.~\ref{fig:selection_16k}), though less consistently than for AutoInterp.

\noindent\textbf{Data Quality Is Objective Dependent.} The matched-budget comparison shows that reducing SAE training data is not characterized by a single notion of data quality. RS better preserves several properties of the full activation stream, most clearly reconstruction, while FPS can improve metrics like AutoInterp and, in some settings of unlearning. Thus, \textbf{which activations are selected matters in addition to how many are selected.} A subset that is representative of the original activation distribution and a subset that maximizes geometric diversity optimize \emph{different} notions of SAE quality.

\looseness=-1\noindent\textbf{Why does FPS produce this tradeoff?} To characterize what each selector retains, we compared the L2-norm distribution of residual-stream activations selected for SAE training by RS vs.\ FPS, with all selectors applied to identical activation pools drawn from the 500M-token stream. RS reproduces the stream's distribution within sampling error at every budget, so its effect is purely a reduction in data quantity. FPS instead concentrates training on a higher-norm region of activation space,
shifting the distribution further right as the subset shrinks from 10\% to 1\% of 500M reference (see Table~\ref{tab:fps_distribution_shift}). 
\begin{table*}[t]
\centering
\small
\setlength{\tabcolsep}{2.5pt}
\renewcommand{\arraystretch}{0.9}
\caption{
\textbf{FPS reshapes the training-data distribution; random selection does not.\textsuperscript{*}} Random selection is distribution-neutral at every budget (KL $\le 0.0003$ bits); FPS-induced divergence grows with subset selectivity. Results for Qwen3-8B and Pythia-160M are in Table~\ref{tab:fps_distribution_shift_appendix} in the appendix
}
\label{tab:fps_distribution_shift}
\begin{tabular}{lccccc}
\toprule
&&&
\multicolumn{3}{c}{\textbf{Selected vs.\ 500M stream}} \\
\cmidrule(lr){4-6}

\textbf{Model}
& \textbf{Selection}
& \textbf{Coreset}
& \multicolumn{1}{c}{$\boldsymbol{\Delta}$\textbf{ mean norm (\%)}}
& \multicolumn{1}{c}{\textbf{KL (bits)}}
& \multicolumn{1}{c}{\textbf{Mean-dir.\ cosine}} \\
\midrule

\multirow{6}{*}{Gemma-2-2B}
& \multirow{3}{*}{FPS}
& 5M  & $+28.8 \pm 0.9$ & 2.28 & $0.941 \pm 0.003$ \\
&& 25M & $+22.7 \pm 0.3$ & 1.51 & $0.954 \pm 0.001$ \\
&& 50M & $+18.9 \pm 0.2$ & 1.11 & $0.961 \pm 0.001$ \\
\cmidrule(lr){2-6}
& \multirow{3}{*}{Random}
& 5M  & $0.0 \pm 0.4$  & 0.000 & $1.000 \pm 0.000$ \\
&& 25M & $0.0 \pm 0.2$  & 0.000 & $1.000 \pm 0.000$ \\
&& 50M & $0.0 \pm 0.1$  & 0.000 & $1.000 \pm 0.000$ \\





\bottomrule
\end{tabular}

\vspace{2pt}
\textsuperscript{*}\emph{\(\Delta\) mean norm} is the relative shift in mean
activation \(L_2\) norm; \emph{KL} is the base-2 divergence between binned norm
distributions; and \emph{Mean-dir.\ cosine} measures mean-direction alignment.
All selectors use identical activation pools. Values are mean \(\pm\) SD across
pool refreshes (50 for norm statistics; 15 for direction statistics); values
shown as \(\pm 0.000\) have SD \(<10^{-4}\).
\vspace{-0.5cm}
\end{table*}

\section{Functional Similarity Does Not Imply Feature Reproducibility}
\label{sec:feature_reproducibility}

So far, increasing activation budget does not uniformly improve SAE function: reconstruction often approaches the \(500\)M reference early, while other metrics remain heterogeneous and non-monotonic. We now ask a stricter question: \emph{does functional similarity imply recovery of the same features?}

Following prior work on SAE feature reproducibility~\cite{paulo2025same}, we
compare reduced-data SAEs against the corresponding \(500\)M-token reference
using one-to-one Hungarian matching. Encoder and decoder features are matched independently by cosine similarity. A latent is \emph{shared} at the threshold \(\tau\) if both matchings select the same reference latent and both cosine similarities exceed \(\tau\). Unless otherwise stated, we use \(\tau=0.7\) as a strict
one-to-one feature-recovery criterion.

\xhdr{Counterpart Agreement Emerges Before Strong Feature Recovery} Even at small activation budgets, reduced-data SAEs are not learning completely unrelated dictionaries. For Gemma-2-2B BatchTopK at \(5\)M tokens, the mean matched decoder cosine similarity is \(0.454\), and independently computed encoder and decoder matchings identify the same reference counterpart for \(66.6\%\) of latents. However, only \(2.20\%\) of latents satisfy the stricter shared-feature criterion at \(\tau=0.7\). 

This gap distinguishes two different notions of correspondence. \emph{Counterpart agreement} asks whether the encoder and decoder identify the same reference feature, whereas \emph{strong feature recovery} additionally requires the corresponding feature directions themselves to be highly aligned. Reduced-data SAEs can therefore recover recognizable counterparts well before those counterparts become strongly reproducible.

\looseness=-1\xhdr{Feature Reproducibility Improves Systematically with Data} 
Table~\ref{tab:feature_correspondence} shows that feature recovery strengthens consistently with activation budget. For Gemma-2-2B BatchTopK, encoder-decoder agreement rises from \(66.61\%\) at \(5\)M to \(83.47\%\) at \(50\)M, while strict shared-feature recovery at \(\tau=0.7\) increases from \(2.20\%\) to \(28.86\%\). The same trend holds for ReLU SAEs, showing that counterpart agreement emerges earlier than strong one-to-one feature recovery. Additional results are in Table~\ref{tab:feature_correspondence_all} in the appendix.

\begin{table}[t]
\centering
\small
\setlength{\tabcolsep}{6pt}
\renewcommand{\arraystretch}{0.95}
\caption{
\textbf{Counterpart agreement and strict feature correspondence.}
\emph{Enc.-dec. agree} is the fraction of features for which independently
computed encoder and decoder Hungarian matchings select the same \(500\)M
reference latent. \emph{Shared} additionally requires both encoder and decoder
cosine similarities to exceed \(\tau=0.7\). We find a similar observation for both Qwen3-8b and Pythia-160m, shown in Table ~\ref{tab:feature_correspondence_all} in the appendix.
}
\label{tab:feature_correspondence}

\begin{tabular}{llccc}
\toprule
\textbf{Model}
& \textbf{SAE}
& \textbf{Budget}
& \textbf{Counterpart (\%)}
& \textbf{Shared, \(\tau\ge0.7\) (\%)} \\
\midrule

\multirow{6}{*}{Gemma-2-2B}
& \multirow{3}{*}{ReLU}
& 5M  & 76.15 & 28.42 \\
&& 25M & 83.06 & 41.13 \\
&& 50M & \textbf{85.28} & \textbf{47.26} \\
\cmidrule(lr){2-5}
& \multirow{3}{*}{BatchTopK}
& 5M  & 66.61 & 2.20 \\
&& 25M & 80.15 & 16.94 \\
&& 50M & \textbf{83.47} & \textbf{28.86} \\


\bottomrule
\end{tabular}
\vspace{-0.5cm}
\end{table}


\xhdr{Qualitative Examples Illustrate the Same Distinction} Table~\ref{tab:qualitative_feature_alignment} presents representative Gemma-2-2B ReLU examples spanning low, medium, and high mean encoder-decoder alignment. For each 5M-500M SAE pair, Hungarian matching aligns features using encoder-decoder cosine similarity, while Gemma-based AutoInterp provides balanced-accuracy scores. We retain pairs for which the encoder- and decoder-based matches agree, and both AutoInterp scores are positive and select one representative pair from each of the alignment bins for qualitative analysis. At low and medium alignment, matched features can have substantially different semantic descriptions despite both matching criteria identifying them as corresponding. In contrast, the highly aligned example captures essentially the same semantic pattern across the two SAEs: \textit{Comparative adjectives and adverbs, of-
ten followed by than}. This qualitative behavior is consistent with the quantitative results above.
Counterpart identity alone is a weak notion of feature recovery; strongly
aligned matched pairs provide a much stricter indication that the same feature
has been recovered across data budgets.

\begin{table*}[t]
\centering
\small
\setlength{\tabcolsep}{2.5pt}
\renewcommand{\arraystretch}{0.9}
\caption{
Representative matched Gemma-2-2B ReLU features between the \(5\)M SAE
and the \(500\)M reference across increasing mean encoder-decoder alignment. The matched latents are ($2782\leftrightarrow2782$), ($411\leftrightarrow2629$),($1507\leftrightarrow1507$).
}
\label{tab:qualitative_feature_alignment}

\begin{tabularx}{\textwidth}{@{}llXX@{}}
\toprule
\textbf{Regime}
& \textbf{Mean alignment}
& \textbf{5M SAE feature}
& \textbf{500M matched feature} \\
\midrule

\textbf{low}
& \(0.0\text{--}0.2\)
& The word \emph{average} and surrounding mathematical or statistical
  expressions
& Mathematical formatting, LaTeX code, variable subscripts, or mathematical
  expressions \\



\textbf{Medium}
& \(0.4\text{--}0.6\)
& Words related to medical topics, particularly cancer, diseases, and research

& Mathematical expressions, code snippets, formatting symbols, or sequential
  lists  \\



\addlinespace[2pt]

\textbf{high}
& \(0.8\text{--}1.0\)
& Comparative adjectives and adverbs, often followed by \emph{than} or
  \emph{as} 
& Comparative adjectives and adverbs, often followed by words such as
  \emph{than} or \emph{more}  \\

\bottomrule
\end{tabularx}
\vspace{-0.4cm}
\end{table*}

\xhdr{Functional Similarity Does Not Imply Feature Reproducibility}
By \(25\)M--\(50\)M tokens, reconstruction and several functional SAE metrics
are already close to their \(500\)M reference values in many regimes, yet
feature recovery remains incomplete. Table~\ref{tab:feature_correspondence}
shows that increasing activation exposure consistently strengthens one-to-one
feature correspondence across models and SAE architectures, even when functional
metrics do not improve monotonically with data. Thus, functional similarity can
emerge at smaller budgets, while feature reproducibility continues to improve
with additional activation exposure, making them distinct notions of SAE
convergence.


\section{Conclusion}

In this work, we ask a simple but practically important question: \emph{How much activation data does an SAE actually need?} We study this across four SAE architectures and three language models, analyzing data requirements along three axes: theory, functional performance, and feature reproducibility. First, we develop a theoretical framework characterizing when reduced-data and full-data SAEs should recover the same feature system, showing that convergence to the same population objective is not sufficient for one-to-one feature recovery; identifiability and optimization conditions are also required. Second, training SAEs on $1\%$, $5\%$, $10\%$, and $100\%$ of the available online activation data, we find that many SAEBench metrics approach the large-data reference well before the full budget. Third, feature-level analysis reveals a different regime: encoder- and decoder-based Hungarian matchings agree for a substantial fraction of features even at small budgets, but stricter one-to-one similarity criteria show continued improvement with more data. Counterpart agreement thus emerges earlier than strong feature reproducibility.

Together, these results suggest that SAE data requirements are objective-dependent. Smaller activation budgets can often preserve reconstruction and aggregate functional behavior, while reproducible recovery of individual features may require substantially more data. We therefore recommend matching SAE training scale to the intended use: avoid defaulting to the largest available activation corpus for functional evaluation, but explicitly assess feature reproducibility when individual latents are used as scientific objects.

\xhdr{Limitations} Our study relies heavily on the training and evaluation implementations provided by SAEBench. All functional metrics are computed using the corresponding SAEBench evaluation scripts, and our SAE training pipeline follows the same underlying framework. Consequently, our conclusions inherit any assumptions,
implementation choices, and metric-specific limitations of these tools. Additionally, the current SAEBench evaluation framework supports RAVEL and unlearning evaluations only for Gemma models. Consequently, we are currently unable to evaluate the Qwen and Pythia models using these metrics.


\bibliographystyle{unsrtnat}
\bibliography{reference}

@String(NeurIPS = {Adv. Neural Inform. Process. Syst.})

@String(ICML  = {Int. Conf. Mach. Learn.})

@String(ICLR  = {Int. Conf. Learn. Represent.})

@String(TMLR  = {Trans. Mach. Learn Res.})

@String(NeurIPS = {NeurIPS})

@String(ICML  = {ICML})

@String(ICLR  = {ICLR})

@String(TMLR  = {TMLR})

@inproceedings{cunningham2023sparse,
  title     = {Sparse Autoencoders Find Highly Interpretable Features in Language Models},
  author    = {Cunningham, Hoagy and Ewart, Aidan and Riggs, Logan and Huben, Robert and Sharkey, Lee},
  booktitle = {International Conference on Learning Representations (ICLR)},
  year      = {2024},
  note      = {arXiv:2309.08600}
}

@misc{marks2024dictionary_learning,
   title = {dictionary\_learning},
   author = {Samuel Marks and Adam Karvonen and Aaron Mueller},
   year = {2024},
   howpublished = {\url{https://github.com/saprmarks/dictionary_learning}},
}

@misc{bricken2023monosemanticity,
  title        = {Towards Monosemanticity: Decomposing Language Models With Dictionary Learning},
  author       = {Bricken, Trenton and Templeton, Adly and Batson, Joshua and Chen, Brian and Jermyn, Adam and Conerly, Tom and Turner, Nick and Anil, Cem and Denison, Carson and Askell, Amanda and Lasenby, Robert and Wu, Yifan and Kravec, Shauna and Schiefer, Nicholas and Maxwell, Tim and Joseph, Nicholas and Hatfield-Dodds, Zac and Tamkin, Alex and Nguyen, Karina and McLean, Brayden and Burke, Josiah E. and Hume, Tristan and Carter, Shan and Henighan, Tom and Olah, Christopher},
  howpublished = {Transformer Circuits Thread},
  year         = {2023},
  url          = {https://transformer-circuits.pub/2023/monosemantic-features/index.html}
}

@misc{templeton2024scaling,
  title        = {Scaling Monosemanticity: Extracting Interpretable Features from {Claude 3 Sonnet}},
  author       = {Templeton, Adly and Conerly, Tom and Marcus, Jonathan and Lindsey, Jack and Bricken, Trenton and others},
  howpublished = {Transformer Circuits Thread},
  year         = {2024},
  url          = {https://transformer-circuits.pub/2024/scaling-monosemanticity/index.html}
}

@inproceedings{gao2024scaling,
  title     = {Scaling and Evaluating Sparse Autoencoders},
  author    = {Gao, Leo and Dupr{\'e} la Tour, Tom and Tillman, Henk and Goh, Gabriel and Troll, Rajan and Radford, Alec and Sutskever, Ilya and Leike, Jan and Wu, Jeffrey},
  booktitle = {International Conference on Learning Representations (ICLR)},
  year      = {2025},
  note      = {arXiv:2406.04093}
}

@article{lieberum2024gemma,
  title   = {{Gemma Scope}: Open Sparse Autoencoders Everywhere All At Once on {Gemma 2}},
  author  = {Lieberum, Tom and Rajamanoharan, Senthooran and Conmy, Arthur and Smith, Lewis and Sonnerat, Nicolas and Varma, Vikrant and Kram{\'a}r, J{\'a}nos and Dragan, Anca and Shah, Rohin and Nanda, Neel},
  journal = {arXiv preprint arXiv:2408.05147},
  year    = {2024}
}

@article{sharkey2025open,
  title   = {Open Problems in Mechanistic Interpretability},
  author  = {Sharkey, Lee and Chughtai, Bilal and Batson, Joshua and Lindsey, Jack and Wu, Jeff and others},
  journal = {arXiv preprint arXiv:2501.16496},
  year    = {2025}
}

@article{olah2020zoom,
  title   = {Zoom In: An Introduction to Circuits},
  author  = {Olah, Chris and Cammarata, Nick and Schubert, Ludwig and Goh, Gabriel and Petrov, Michael and Carter, Shan},
  journal = {Distill},
  year    = {2020}
}

@misc{elhage2022superposition,
  title        = {Toy Models of Superposition},
  author       = {Elhage, Nelson and Hume, Tristan and Olsson, Catherine and Schiefer, Nicholas and Henighan, Tom and others},
  howpublished = {Transformer Circuits Thread},
  year         = {2022},
  url          = {https://transformer-circuits.pub/2022/toy_model/index.html}
}

@article{olshausen1996emergence,
  title   = {Emergence of Simple-Cell Receptive Field Properties by Learning a Sparse Code for Natural Images},
  author  = {Olshausen, Bruno A. and Field, David J.},
  journal = {Nature},
  volume  = {381},
  number  = {6583},
  pages   = {607--609},
  year    = {1996}
}

@article{rajamanoharan2024jumping,
  title   = {Jumping Ahead: Improving Reconstruction Fidelity with {JumpReLU} Sparse Autoencoders},
  author  = {Rajamanoharan, Senthooran and Lieberum, Tom and Sonnerat, Nicolas and Conmy, Arthur and Varma, Vikrant and Kram{\'a}r, J{\'a}nos and Nanda, Neel},
  journal = {arXiv preprint arXiv:2407.14435},
  year    = {2024}
}

@inproceedings{bussmann2024batchtopk,
  title     = {{BatchTopK} Sparse Autoencoders},
  author    = {Bussmann, Bart and Leask, Patrick and Nanda, Neel},
  booktitle = {NeurIPS 2024 Workshop on Scientific Methods for Understanding Deep Learning},
  year      = {2024}
}

@inproceedings{bussmann2025matryoshka,
  title     = {Learning Multi-Level Features with Matryoshka Sparse Autoencoders},
  author    = {Bussmann, Bart and Nabeshima, Noa and Karvonen, Adam and Nanda, Neel},
  booktitle = {International Conference on Machine Learning (ICML)},
  year      = {2025}
}

@inproceedings{karvonen2025saebench,
  title     = {{SAEBench}: A Comprehensive Benchmark for Sparse Autoencoders in Language Model Interpretability},
  author    = {Karvonen, Adam and Rager, Can and Lin, Johnny and Tigges, Curt and Bloom, Joseph and Chanin, David and Lau, Yeu-Tong and Farrell, Eoin and McDougall, Callum and Ayonrinde, Kola and Till, Demian and Wearden, Matthew and Conmy, Arthur and Marks, Samuel and Nanda, Neel},
  booktitle = {International Conference on Machine Learning (ICML)},
  year      = {2025}
}

@inproceedings{chanin2024absorption,
  title     = {A is for Absorption: Studying Feature Splitting and Absorption in Sparse Autoencoders},
  author    = {Chanin, David and Wilken-Smith, James and Dulka, Tom{\'a}{\v{s}} and Bhatnagar, Hardik and Golechha, Satvik and Bloom, Joseph},
  booktitle = {Advances in Neural Information Processing Systems (NeurIPS)},
  year      = {2025}
}

@misc{bills2023language,
  title        = {Language Models Can Explain Neurons in Language Models},
  author       = {Bills, Steven and Cammarata, Nick and Mossing, Dan and Tillman, Henk and Gao, Leo and Goh, Gabriel and Sutskever, Ilya and Leike, Jan and Wu, Jeff and Saunders, William},
  howpublished = {OpenAI},
  year         = {2023},
  url          = {https://openaipublic.blob.core.windows.net/neuron-explainer/paper/index.html}
}

@inproceedings{paulo2024automatically,
  title     = {Automatically Interpreting Millions of Features in Large Language Models},
  author    = {Paulo, Gon{\c{c}}alo and Mallen, Alex and Juang, Caden and Belrose, Nora},
  booktitle = {International Conference on Machine Learning (ICML)},
  year      = {2025}
}

@article{gurnee2023finding,
  title   = {Finding Neurons in a Haystack: Case Studies with Sparse Probing},
  author  = {Gurnee, Wes and Nanda, Neel and Pauly, Matthew and Harvey, Katherine and Troitskii, Dmitrii and Bertsimas, Dimitris},
  journal = {Transactions on Machine Learning Research (TMLR)},
  year    = {2023}
}

@inproceedings{marks2024sparse,
  title     = {Sparse Feature Circuits: Discovering and Editing Interpretable Causal Graphs in Language Models},
  author    = {Marks, Samuel and Rager, Can and Michaud, Eric J. and Belinkov, Yonatan and Bau, David and Mueller, Aaron},
  booktitle = {International Conference on Learning Representations (ICLR)},
  year      = {2025},
  note      = {arXiv:2403.19647}
}

@inproceedings{huang2024ravel,
  title     = {{RAVEL}: Evaluating Interpretability Methods on Disentangling Language Model Representations},
  author    = {Huang, Jing and Wu, Zhengxuan and Potts, Christopher and Geva, Mor and Geiger, Atticus},
  booktitle = {Proceedings of the 62nd Annual Meeting of the Association for Computational Linguistics (ACL)},
  year      = {2024}
}

@article{farrell2024applying,
  title   = {Applying Sparse Autoencoders to Unlearn Knowledge in Language Models},
  author  = {Farrell, Eoin and Lau, Yeu-Tong and Conmy, Arthur},
  journal = {arXiv preprint arXiv:2410.19278},
  year    = {2024}
}

@inproceedings{paulo2025same,
  title     = {Sparse Autoencoders Trained on the Same Data Learn Different Features},
  author    = {Paulo, Gon{\c{c}}alo and Belrose, Nora},
  booktitle = {International Conference on Learning Representations (ICLR)},
  year      = {2026}
}

@article{lan2024universality,
  title   = {Quantifying Feature Space Universality Across Large Language Models via Sparse Autoencoders},
  author  = {Lan, Michael and Torr, Philip and Meek, Austin and Khakzar, Ashkan and Krueger, David and Barez, Fazl},
  journal = {arXiv preprint arXiv:2410.06981},
  year    = {2024}
}

@article{gerasimov2026unstable,
  title   = {Unstable Features, Reproducible Subspaces: Understanding Seed Dependence in Sparse Autoencoders},
  author  = {Gerasimov, Gleb and Rusalev, Timofei and Balagansky, Nikita and Laptev, Daniil and Kurochkin, Vadim and Gavrilov, Daniil},
  journal = {arXiv preprint arXiv:2606.12138},
  year    = {2026}
}

@article{kuhn1955hungarian,
  title   = {The {H}ungarian Method for the Assignment Problem},
  author  = {Kuhn, Harold W.},
  journal = {Naval Research Logistics Quarterly},
  volume  = {2},
  number  = {1--2},
  pages   = {83--97},
  year    = {1955}
}

@article{moser2025coreset,
  title   = {A Coreset Selection of Coreset Selection Literature: Introduction and Recent Advances},
  author  = {Moser, Brian B. and Shanbhag, Arundhati S. and Frolov, Stanislav and Raue, Federico and Folz, Joachim and Dengel, Andreas},
  journal = {arXiv preprint arXiv:2505.17799},
  year    = {2025}
}

@inproceedings{coleman2020selection,
  title     = {Selection via Proxy: Efficient Data Selection for Deep Learning},
  author    = {Coleman, Cody and Yeh, Christopher and Mussmann, Stephen and Mirzasoleiman, Baharan and Bailis, Peter and Liang, Percy and Leskovec, Jure and Zaharia, Matei},
  booktitle = {International Conference on Learning Representations (ICLR)},
  year      = {2020},
  note      = {arXiv:1906.11829}
}

@article{yang2025diversity,
  title   = {Diversity-driven Data Selection for Language Model Tuning through Sparse Autoencoder},
  author  = {Yang, Xianjun and Nie, Shaoliang and Liu, Lijuan and Gururangan, Suchin and Karn, Ujjwal and Hou, Rui and Khabsa, Madian and Mao, Yuning},
  journal = {arXiv preprint arXiv:2502.14050},
  year    = {2025}
}

@article{gonzalez1985clustering,
  title   = {Clustering to Minimize the Maximum Intercluster Distance},
  author  = {Gonzalez, Teofilo F.},
  journal = {Theoretical Computer Science},
  volume  = {38},
  pages   = {293--306},
  year    = {1985}
}

@article{eldar1997farthest,
  title   = {The Farthest Point Strategy for Progressive Image Sampling},
  author  = {Eldar, Yuval and Lindenbaum, Michael and Porat, Moshe and Zeevi, Yehoshua Y.},
  journal = {IEEE Transactions on Image Processing},
  volume  = {6},
  number  = {9},
  pages   = {1305--1315},
  year    = {1997}
}

@inproceedings{sener2018active,
  title     = {Active Learning for Convolutional Neural Networks: A Core-Set Approach},
  author    = {Sener, Ozan and Savarese, Silvio},
  booktitle = {International Conference on Learning Representations (ICLR)},
  year      = {2018}
}

@inproceedings{biderman2023pythia,
  title     = {Pythia: A Suite for Analyzing Large Language Models Across Training and Scaling},
  author    = {Biderman, Stella and Schoelkopf, Hailey and Anthony, Quentin Gregory and Bradley, Herbie and O'Brien, Kyle and Hallahan, Eric and Khan, Mohammad Aflah and Purohit, Shivanshu and Prashanth, USVSN Sai and Raff, Edward and Skowron, Aviya and Sutawika, Lintang and van der Wal, Oskar},
  booktitle = {International Conference on Machine Learning (ICML)},
  year      = {2023},
  note      = {arXiv:2304.01373}
}

@article{gemma2024gemma2,
  title   = {{Gemma 2}: Improving Open Language Models at a Practical Size},
  author  = {{Gemma Team}},
  journal = {arXiv preprint arXiv:2408.00118},
  year    = {2024}
}

@article{yang2025qwen3,
  title   = {{Qwen3} Technical Report},
  author  = {Yang, An and Li, Anfeng and Yang, Baosong and Zhang, Beichen and Hui, Binyuan and others},
  journal = {arXiv preprint arXiv:2505.09388},
  year    = {2025}
}

@inproceedings{penedo2024fineweb,
  title     = {The {FineWeb} Datasets: Decanting the Web for the Finest Text Data at Scale},
  author    = {Penedo, Guilherme and Kydl{\'i}{\v{c}}ek, Hynek and Ben Allal, Loubna and Lozhkov, Anton and Mitchell, Margaret and Raffel, Colin and Von Werra, Leandro and Wolf, Thomas},
  booktitle = {Advances in Neural Information Processing Systems (NeurIPS) Datasets and Benchmarks Track},
  year      = {2024},
  note      = {arXiv:2406.17557}
}
\appendix
\appendix

\section{Experimental details and additional Functional Data-Scaling Results}
\label{appendix:functional_saturation}

We evaluate four SAE architectures: standard ReLU, BatchTopK, JumpReLU, and Matryoshka BatchTopK. Across all configurations, SAE batch size \(2048\), dictionary widths \(16{,}384\) and
\(65{,}536\), \(1000\) warmup steps, and \(5000\) sparsity-warmup steps. Decay begins after \(80\%\) of training, TopK-style \(k\)-annealing ends after \(1\%\) of the total training steps, BOS tokens are removed, and activations exceeding \(10\times\) the reference activation norm are filtered. The default
full-data training budget is \(500\)M activation tokens. For activation extraction, Pythia-160M uses LM batch size \(32\), context length \(1024\), and \texttt{float32}; Gemma-2-2B uses batch size \(4\), context length \(1024\), and \texttt{bfloat16}; and Qwen3-8B uses batch size \(16\), context length
\(2048\), and \texttt{bfloat16}. For Gemma-2-2B and Pythia-160M, we evaluate both 16K and 65K dictionary widths. For Qwen3-8B, we use only the 65K width across all SAE architectures.

The trainer index used in our evaluation tables identifies the
architecture-specific sparsity operating point. We use the
\(L_1\) sparsity penalty over$
\{0.012,\,0.015,\,0.020,\,0.030,\,0.040,\,0.060\}.
$
At dictionary width \(16\)K, trainers \(0\)-\(5\) correspond to these values
in order; at width \(64\)K, trainers \(6\)-\(11\) follow the same ordering.

For BatchTopK, we use \(k\in\{80,160\}\). Trainers \(0\) and \(1\) correspond to \(k=80\) and
\(k=160\) at width \(16\)K, while trainers \(2\) and \(3\) correspond to the
same two operating points at width \(64\)K. The trainer uses
\(\texttt{auxk\_alpha}=1/32\), \(\texttt{threshold\_beta}=0.999\), and begins
threshold tracking after \(1000\) steps.

For JumpReLU, we use the target activation count
$L_0^{\mathrm{target}}\in\{80,160\}$.
Trainers \(0\) and \(1\) correspond to target \(L_0=80\) and \(160\) at width
\(16\)K, while trainers \(2\) and \(3\) correspond to the same targets at width
\(64\)K. The sparsity penalty is fixed to \(1.0\) and the JumpReLU bandwidth
to \(0.001\).

For Matryoshka BatchTopK, we use \(k\in\{80,160\}\).
Trainers \(0\) and \(1\) correspond to \(k=80\) and \(160\) at width \(16\)K,
and trainers \(2\) and \(3\) correspond to the same values at width \(64\)K.
The nested groups use fractions
$
\left\{
\frac{1}{32},\,
\frac{1}{16},\,
\frac{1}{8},\,
\frac{1}{4},\,
\frac{17}{32}
\right\},
$
with \(\texttt{auxk\_alpha}=1/32\),
\(\texttt{threshold\_beta}=0.999\), and threshold tracking beginning after
\(1000\) steps.

\begin{table*}[t]
\centering
\small
\setlength{\tabcolsep}{3.5pt}
\renewcommand{\arraystretch}{1.08}
\caption{\textbf{\textbf{SAE evaluation across activation budgets and selection strategies for Gemma-2-2B, ReLU (16k). Random and FPS are compared at 5M,25M, and 50M activation budgets against the corresponding 500M baseline.}}}
\label{tab:pertrainer_gemma_standard_new_16384}
\resizebox{\textwidth}{!}{
\begin{tabular}{lllccccccccc}
\toprule
\textbf{Trainer} & \textbf{Budget} & \textbf{Selection} & \shortstack{\textbf{Loss}\\\textbf{Recovered}\\$\uparrow$} & \shortstack{\textbf{$1-$Abs.}\\\textbf{Fraction}\\$\uparrow$} & \shortstack{\textbf{AutoInterp}\\$\uparrow$} & \shortstack{\textbf{SCR}\\\textbf{Top-20}\\$\uparrow$} & \shortstack{\textbf{TPP}\\\textbf{Top-20}\\$\uparrow$} & \shortstack{\textbf{Sparse Probe}\\\textbf{Top-1}\\$\uparrow$} & \shortstack{\textbf{RAVEL}\\\textbf{Disent.}\\$\uparrow$} & \shortstack{\textbf{Unlearning}\\$\uparrow$} & \shortstack{\textbf{Sparsity}\\\textbf{(L0)}\\$\downarrow$} \\
\midrule
  \multirow{7}{*}{0} & \multirow{2}{*}{5M} & Random & 0.993 & 0.959 & 0.703 & 0.253 & 0.029 & 0.756 & \textbf{0.657} & 0.006 & 1400.2 \\
   &  & FPS & 0.985 & 0.974 & 0.703 & 0.194 & 0.021 & 0.750 & 0.639 & \textbf{0.009} & 1298.8 \\
  \cmidrule(lr){2-12}
   & \multirow{2}{*}{25M} & Random & \textbf{0.995} & 0.974 & 0.704 & 0.261 & \textbf{0.031} & 0.787 & 0.653 & 0.002 & 1353.2 \\
   &  & FPS & 0.992 & \textbf{0.983} & \textbf{0.718} & 0.223 & 0.022 & 0.802 & 0.634 & 0.004 & 1265.1 \\
  \cmidrule(lr){2-12}
   & \multirow{2}{*}{50M} & Random & \textbf{0.995} & 0.979 & 0.715 & 0.264 & 0.030 & 0.797 & 0.656 & \textbf{0.009} & 1346.5 \\
   &  & FPS & 0.993 & 0.955 & 0.712 & 0.230 & 0.024 & 0.772 & 0.625 & 0.004 & \textbf{1204.0} \\
  \cmidrule(lr){2-12}
   & \multirow{1}{*}{500M} & Baseline & \textbf{0.995} & 0.947 & 0.712 & \textbf{0.268} & 0.030 & \textbf{0.805} & 0.641 & \textbf{0.009} & 1289.5 \\
\bottomrule
\end{tabular}
}
\end{table*}

\begin{table*}[t]
\centering
\small
\setlength{\tabcolsep}{3.5pt}
\renewcommand{\arraystretch}{1.08}
\caption{\textbf{Per-trainer SAE evaluation — Gemma-2-2B, ReLU (64K).} Each row is one trainer run. Missing evaluations shown as ``--''.}
\label{tab:pertrainer_gemma_standard_new_65536}
\resizebox{\textwidth}{!}{
\begin{tabular}{lllccccccccc}
\toprule
\textbf{Trainer} & \textbf{Budget} & \textbf{Selection} & \shortstack{\textbf{Loss}\\\textbf{Recovered}\\$\uparrow$} & \shortstack{\textbf{$1-$Abs.}\\\textbf{Fraction}\\$\uparrow$} & \shortstack{\textbf{AutoInterp}\\$\uparrow$} & \shortstack{\textbf{SCR}\\\textbf{Top-20}\\$\uparrow$} & \shortstack{\textbf{TPP}\\\textbf{Top-20}\\$\uparrow$} & \shortstack{\textbf{Sparse Probe}\\\textbf{Top-1}\\$\uparrow$} & \shortstack{\textbf{RAVEL}\\\textbf{Disent.}\\$\uparrow$} & \shortstack{\textbf{Unlearning}\\$\uparrow$} & \shortstack{\textbf{Sparsity}\\\textbf{(L0)}\\$\downarrow$} \\
\midrule
  \multirow{7}{*}{6} & \multirow{2}{*}{5M} & Random & 0.995 & \textbf{0.902} & 0.648 & 0.191 & 0.011 & 0.775 & \textbf{0.690} & 0.024 & 1700.5 \\
   &  & FPS & 0.988 & 0.894 & 0.663 & 0.125 & 0.008 & 0.792 & 0.664 & 0.013 & \textbf{1439.5} \\
  \cmidrule(lr){2-12}
   & \multirow{2}{*}{25M} & Random & 0.995 & 0.865 & 0.669 & \textbf{0.192} & \textbf{0.012} & 0.811 & 0.685 & 0.011 & 1932.7 \\
   &  & FPS & 0.993 & 0.758 & 0.695 & 0.155 & 0.009 & 0.805 & 0.665 & 0.009 & 1686.4 \\
  \cmidrule(lr){2-12}
   & \multirow{2}{*}{50M} & Random & \textbf{0.997} & 0.819 & 0.673 & 0.183 & 0.012 & \textbf{0.813} & 0.685 & 0.021 & 1987.0 \\
   &  & FPS & 0.995 & 0.729 & \textbf{0.702} & 0.159 & 0.009 & 0.812 & 0.681 & \textbf{0.051} & 1678.4 \\
  \cmidrule(lr){2-12}
   & \multirow{1}{*}{500M} & Baseline & \textbf{0.997} & 0.788 & 0.689 & 0.184 & 0.012 & 0.807 & 0.683 & 0.006 & 1923.7 \\
\bottomrule
\end{tabular}
}
\end{table*}

\begin{table*}[t]
\centering
\small
\setlength{\tabcolsep}{3.5pt}
\renewcommand{\arraystretch}{1.08}
\caption{\textbf{Per-trainer SAE evaluation for Gemma-2-2B, BatchTopK (16K).}}
\label{tab:pertrainer_gemma_batch_top_k_16384}
\resizebox{\textwidth}{!}{
\begin{tabular}{lllccccccccc}
\toprule
\textbf{Trainer} & \textbf{Budget} & \textbf{Selection} & \shortstack{\textbf{Loss}\\\textbf{Recovered}\\$\uparrow$} & \shortstack{\textbf{$1-$Abs.}\\\textbf{Fraction}\\$\uparrow$} & \shortstack{\textbf{AutoInterp}\\$\uparrow$} & \shortstack{\textbf{SCR}\\\textbf{Top-20}\\$\uparrow$} & \shortstack{\textbf{TPP}\\\textbf{Top-20}\\$\uparrow$} & \shortstack{\textbf{Sparse Probe}\\\textbf{Top-1}\\$\uparrow$} & \shortstack{\textbf{RAVEL}\\\textbf{Disent.}\\$\uparrow$} & \shortstack{\textbf{Unlearning}\\$\uparrow$} & \shortstack{\textbf{Sparsity}\\\textbf{(L0)}\\$\downarrow$} \\
\midrule
  \multirow{7}{*}{0} & \multirow{2}{*}{5M} & Random & 0.979 & 0.947 & 0.815 & 0.352 & 0.047 & 0.732 & 0.555 & 0.041 & 79.8 \\
   &  & FPS & 0.951 & 0.928 & 0.819 & 0.243 & 0.025 & 0.733 & 0.517 & 0.063 & \textbf{50.9} \\
  \cmidrule(lr){2-12}
   & \multirow{2}{*}{25M} & Random & 0.985 & 0.928 & 0.833 & \textbf{0.366} & 0.043 & \textbf{0.787} & 0.540 & \textbf{0.101} & 80.6 \\
   &  & FPS & 0.974 & 0.891 & 0.845 & 0.268 & 0.025 & 0.724 & 0.535 & 0.024 & 55.2 \\
  \cmidrule(lr){2-12}
   & \multirow{2}{*}{50M} & Random & 0.985 & \textbf{0.950} & 0.835 & 0.356 & \textbf{0.050} & 0.729 & \textbf{0.578} & 0.024 & 79.7 \\
   &  & FPS & 0.979 & 0.844 & \textbf{0.853} & 0.285 & 0.032 & 0.739 & 0.573 & 0.045 & 57.6 \\
  \cmidrule(lr){2-12}
   & \multirow{1}{*}{500M} & Baseline & \textbf{0.987} & 0.907 & 0.821 & 0.354 & 0.039 & 0.765 & 0.573 & 0.065 & 80.2 \\
\bottomrule
\end{tabular}
}
\end{table*}

\begin{table*}[t]
\centering
\small
\setlength{\tabcolsep}{3.5pt}
\renewcommand{\arraystretch}{1.08}
\caption{\textbf{Per-trainer SAE evaluation for Gemma-2-2B, BatchTopK (64K).}}
\label{tab:pertrainer_gemma_batch_top_k_65536}
\resizebox{\textwidth}{!}{
\begin{tabular}{lllccccccccc}
\toprule
\textbf{Trainer} & \textbf{Budget} & \textbf{Selection} & \shortstack{\textbf{Loss}\\\textbf{Recovered}\\$\uparrow$} & \shortstack{\textbf{$1-$Abs.}\\\textbf{Fraction}\\$\uparrow$} & \shortstack{\textbf{AutoInterp}\\$\uparrow$} & \shortstack{\textbf{SCR}\\\textbf{Top-20}\\$\uparrow$} & \shortstack{\textbf{TPP}\\\textbf{Top-20}\\$\uparrow$} & \shortstack{\textbf{Sparse Probe}\\\textbf{Top-1}\\$\uparrow$} & \shortstack{\textbf{RAVEL}\\\textbf{Disent.}\\$\uparrow$} & \shortstack{\textbf{Unlearning}\\$\uparrow$} & \shortstack{\textbf{Sparsity}\\\textbf{(L0)}\\$\downarrow$} \\
\midrule
  \multirow{7}{*}{2} & \multirow{2}{*}{5M} & Random & 0.980 & 0.899 & 0.785 & 0.330 & \textbf{0.052} & \textbf{0.793} & 0.556 & 0.048 & 79.6 \\
   &  & FPS & 0.954 & \textbf{0.900} & 0.848 & 0.255 & 0.022 & 0.769 & 0.536 & 0.095 & \textbf{53.4} \\
  \cmidrule(lr){2-12}
   & \multirow{2}{*}{25M} & Random & 0.987 & 0.896 & 0.826 & 0.327 & 0.038 & 0.789 & \textbf{0.589} & 0.056 & 80.7 \\
   &  & FPS & 0.977 & 0.666 & \textbf{0.863} & 0.231 & 0.028 & 0.721 & 0.585 & 0.102 & 57.4 \\
  \cmidrule(lr){2-12}
   & \multirow{2}{*}{50M} & Random & 0.987 & 0.883 & 0.838 & \textbf{0.339} & 0.028 & 0.739 & 0.566 & 0.102 & 80.6 \\
   &  & FPS & 0.984 & 0.660 & 0.857 & 0.242 & 0.019 & 0.738 & 0.576 & 0.091 & 59.9 \\
  \cmidrule(lr){2-12}
   & \multirow{1}{*}{500M} & Baseline & \textbf{0.990} & 0.750 & 0.831 & 0.261 & 0.019 & 0.731 & 0.584 & \textbf{0.168} & 80.8 \\
\bottomrule
\end{tabular}
}
\end{table*}

\begin{table*}[t]
\centering
\small
\setlength{\tabcolsep}{3.5pt}
\renewcommand{\arraystretch}{1.08}
\caption{\textbf{Per-trainer SAE evaluation for Gemma-2-2B, JumpReLU (16K).}}
\label{tab:pertrainer_gemma_jump_relu_16384}
\resizebox{\textwidth}{!}{
\begin{tabular}{lllccccccccc}
\toprule
\textbf{Trainer} & \textbf{Budget} & \textbf{Selection} & \shortstack{\textbf{Loss}\\\textbf{Recovered}\\$\uparrow$} & \shortstack{\textbf{$1-$Abs.}\\\textbf{Fraction}\\$\uparrow$} & \shortstack{\textbf{AutoInterp}\\$\uparrow$} & \shortstack{\textbf{SCR}\\\textbf{Top-20}\\$\uparrow$} & \shortstack{\textbf{TPP}\\\textbf{Top-20}\\$\uparrow$} & \shortstack{\textbf{Sparse Probe}\\\textbf{Top-1}\\$\uparrow$} & \shortstack{\textbf{RAVEL}\\\textbf{Disent.}\\$\uparrow$} & \shortstack{\textbf{Unlearning}\\$\uparrow$} & \shortstack{\textbf{Sparsity}\\\textbf{(L0)}\\$\downarrow$} \\
\midrule
  \multirow{7}{*}{0} & \multirow{2}{*}{5M} & Random & 0.977 & 0.910 & 0.825 & 0.365 & 0.047 & 0.736 & 0.523 & 0.015 & 83.4 \\
   &  & FPS & 0.949 & 0.910 & 0.828 & 0.232 & 0.034 & 0.695 & 0.511 & \textbf{0.094} & \textbf{54.3} \\
  \cmidrule(lr){2-12}
   & \multirow{2}{*}{25M} & Random & 0.982 & 0.895 & 0.811 & \textbf{0.383} & 0.043 & 0.715 & \textbf{0.577} & 0.043 & 83.0 \\
   &  & FPS & 0.972 & 0.842 & 0.844 & 0.290 & 0.029 & 0.691 & 0.542 & 0.079 & 57.3 \\
  \cmidrule(lr){2-12}
   & \multirow{2}{*}{50M} & Random & 0.985 & \textbf{0.914} & 0.813 & 0.374 & \textbf{0.053} & 0.733 & 0.573 & 0.030 & 82.7 \\
   &  & FPS & 0.979 & 0.802 & \textbf{0.849} & 0.277 & 0.031 & 0.729 & 0.567 & 0.026 & 60.3 \\
  \cmidrule(lr){2-12}
   & \multirow{1}{*}{500M} & Baseline & \textbf{0.987} & 0.865 & 0.822 & 0.325 & 0.045 & \textbf{0.748} & 0.572 & 0.090 & 83.0 \\
\bottomrule
\end{tabular}
}
\end{table*}

\begin{table*}[t]
\centering
\small
\setlength{\tabcolsep}{3.5pt}
\renewcommand{\arraystretch}{1.08}
\caption{\textbf{Per-trainer SAE evaluation for Gemma-2-2B, JumpReLU (64K).}}
\label{tab:pertrainer_gemma_jump_relu_65536}
\resizebox{\textwidth}{!}{
\begin{tabular}{lllccccccccc}
\toprule
\textbf{Trainer} & \textbf{Budget} & \textbf{Selection} & \shortstack{\textbf{Loss}\\\textbf{Recovered}\\$\uparrow$} & \shortstack{\textbf{$1-$Abs.}\\\textbf{Fraction}\\$\uparrow$} & \shortstack{\textbf{AutoInterp}\\$\uparrow$} & \shortstack{\textbf{SCR}\\\textbf{Top-20}\\$\uparrow$} & \shortstack{\textbf{TPP}\\\textbf{Top-20}\\$\uparrow$} & \shortstack{\textbf{Sparse Probe}\\\textbf{Top-1}\\$\uparrow$} & \shortstack{\textbf{RAVEL}\\\textbf{Disent.}\\$\uparrow$} & \shortstack{\textbf{Unlearning}\\$\uparrow$} & \shortstack{\textbf{Sparsity}\\\textbf{(L0)}\\$\downarrow$} \\
\midrule
  \multirow{7}{*}{2} & \multirow{2}{*}{5M} & Random & 0.977 & \textbf{0.873} & 0.809 & \textbf{0.369} & \textbf{0.044} & 0.733 & 0.561 & 0.069 & 83.5 \\
   &  & FPS & 0.956 & 0.757 & 0.835 & 0.196 & 0.014 & 0.694 & 0.539 & 0.071 & \textbf{55.5} \\
  \cmidrule(lr){2-12}
   & \multirow{2}{*}{25M} & Random & 0.987 & 0.827 & 0.835 & 0.328 & 0.029 & \textbf{0.765} & 0.567 & 0.081 & 82.9 \\
   &  & FPS & 0.975 & 0.717 & \textbf{0.867} & 0.246 & 0.027 & 0.725 & 0.581 & 0.092 & 59.0 \\
  \cmidrule(lr){2-12}
   & \multirow{2}{*}{50M} & Random & 0.987 & 0.834 & 0.825 & 0.308 & 0.039 & 0.737 & 0.572 & 0.084 & 82.8 \\
   &  & FPS & 0.984 & 0.663 & 0.852 & 0.264 & 0.020 & 0.741 & 0.582 & \textbf{0.221} & 62.0 \\
  \cmidrule(lr){2-12}
   & \multirow{1}{*}{500M} & Baseline & \textbf{0.990} & 0.729 & 0.813 & 0.255 & 0.022 & 0.718 & \textbf{0.582} & 0.133 & 83.0 \\
\bottomrule
\end{tabular}
}
\end{table*}

\begin{table*}[t]
\centering
\small
\setlength{\tabcolsep}{3.5pt}
\renewcommand{\arraystretch}{1.08}
\caption{\textbf{Per-trainer SAE evaluation for Gemma-2-2B, Matryoshka (16K).}}
\label{tab:pertrainer_gemma_matryoshka_batch_top_k_16384}
\resizebox{\textwidth}{!}{
\begin{tabular}{lllccccccccc}
\toprule
\textbf{Trainer} & \textbf{Budget} & \textbf{Selection} & \shortstack{\textbf{Loss}\\\textbf{Recovered}\\$\uparrow$} & \shortstack{\textbf{$1-$Abs.}\\\textbf{Fraction}\\$\uparrow$} & \shortstack{\textbf{AutoInterp}\\$\uparrow$} & \shortstack{\textbf{SCR}\\\textbf{Top-20}\\$\uparrow$} & \shortstack{\textbf{TPP}\\\textbf{Top-20}\\$\uparrow$} & \shortstack{\textbf{Sparse Probe}\\\textbf{Top-1}\\$\uparrow$} & \shortstack{\textbf{RAVEL}\\\textbf{Disent.}\\$\uparrow$} & \shortstack{\textbf{Unlearning}\\$\uparrow$} & \shortstack{\textbf{Sparsity}\\\textbf{(L0)}\\$\downarrow$} \\
\midrule
  \multirow{7}{*}{0} & \multirow{2}{*}{5M} & Random & 0.979 & 0.948 & 0.845 & 0.404 & 0.107 & 0.755 & 0.561 & 0.028 & 79.5 \\
   &  & FPS & 0.941 & 0.967 & 0.863 & 0.295 & 0.053 & 0.724 & 0.526 & 0.015 & \textbf{49.1} \\
  \cmidrule(lr){2-12}
   & \multirow{2}{*}{25M} & Random & 0.984 & 0.966 & 0.844 & 0.455 & 0.104 & 0.743 & 0.549 & 0.006 & 80.0 \\
   &  & FPS & 0.969 & 0.953 & \textbf{0.884} & 0.340 & 0.092 & 0.760 & 0.545 & \textbf{0.090} & 54.4 \\
  \cmidrule(lr){2-12}
   & \multirow{2}{*}{50M} & Random & 0.984 & 0.982 & 0.851 & \textbf{0.460} & \textbf{0.130} & 0.776 & 0.562 & 0.041 & 79.5 \\
   &  & FPS & 0.975 & 0.975 & 0.863 & 0.373 & 0.106 & 0.734 & \textbf{0.569} & 0.056 & 57.4 \\
  \cmidrule(lr){2-12}
   & \multirow{1}{*}{500M} & Baseline & \textbf{0.985} & \textbf{0.993} & 0.849 & 0.459 & 0.130 & \textbf{0.806} & 0.560 & 0.034 & 80.1 \\
\bottomrule
\end{tabular}
}
\end{table*}

\begin{table*}[t]
\centering
\small
\setlength{\tabcolsep}{3.5pt}
\renewcommand{\arraystretch}{1.08}
\caption{\textbf{Per-trainer SAE evaluation for Gemma-2-2B, Matryoshka (64K).}}
\label{tab:pertrainer_gemma_matryoshka_batch_top_k_65536}
\resizebox{\textwidth}{!}{
\begin{tabular}{lllccccccccc}
\toprule
\textbf{Trainer} & \textbf{Budget} & \textbf{Selection} & \shortstack{\textbf{Loss}\\\textbf{Recovered}\\$\uparrow$} & \shortstack{\textbf{$1-$Abs.}\\\textbf{Fraction}\\$\uparrow$} & \shortstack{\textbf{AutoInterp}\\$\uparrow$} & \shortstack{\textbf{SCR}\\\textbf{Top-20}\\$\uparrow$} & \shortstack{\textbf{TPP}\\\textbf{Top-20}\\$\uparrow$} & \shortstack{\textbf{Sparse Probe}\\\textbf{Top-1}\\$\uparrow$} & \shortstack{\textbf{RAVEL}\\\textbf{Disent.}\\$\uparrow$} & \shortstack{\textbf{Unlearning}\\$\uparrow$} & \shortstack{\textbf{Sparsity}\\\textbf{(L0)}\\$\downarrow$} \\
\midrule
  \multirow{7}{*}{2} & \multirow{2}{*}{5M} & Random & 0.975 & 0.925 & 0.772 & 0.399 & 0.086 & 0.756 & 0.543 & 0.026 & 79.5 \\
   &  & FPS & 0.949 & 0.967 & 0.873 & 0.266 & 0.074 & 0.760 & 0.540 & \textbf{0.086} & \textbf{50.6} \\
  \cmidrule(lr){2-12}
   & \multirow{2}{*}{25M} & Random & 0.985 & \textbf{0.992} & 0.832 & 0.393 & \textbf{0.106} & \textbf{0.783} & 0.602 & 0.017 & 80.1 \\
   &  & FPS & 0.975 & 0.958 & \textbf{0.885} & 0.339 & 0.048 & 0.756 & \textbf{0.612} & 0.038 & 55.1 \\
  \cmidrule(lr){2-12}
   & \multirow{2}{*}{50M} & Random & \textbf{0.987} & 0.974 & 0.829 & 0.428 & 0.101 & 0.767 & 0.610 & 0.043 & 79.9 \\
   &  & FPS & 0.980 & 0.961 & 0.881 & 0.325 & 0.086 & 0.772 & 0.581 & 0.079 & 57.9 \\
  \cmidrule(lr){2-12}
   & \multirow{1}{*}{500M} & Baseline & \textbf{0.987} & 0.980 & 0.820 & \textbf{0.433} & 0.091 & 0.782 & 0.607 & 0.062 & 79.6 \\
\bottomrule
\end{tabular}
}
\end{table*}


\begin{table*}[t]
\centering
\small
\setlength{\tabcolsep}{3.5pt}
\renewcommand{\arraystretch}{1.08}
\caption{\textbf{Per-trainer SAE evaluation for Pythia-160M, ReLU (16K).}}
\label{tab:pertrainer_pythia_standard_new_16384}
\resizebox{\textwidth}{!}{
\begin{tabular}{lllcccccccc}
\toprule
\textbf{Trainer} & \textbf{Budget} & \textbf{Selection} & \shortstack{\textbf{Loss}\\\textbf{Recovered}\\$\uparrow$} & \shortstack{\textbf{$1-$Abs.}\\\textbf{Fraction}\\$\uparrow$} & \shortstack{\textbf{AutoInterp}\\$\uparrow$} & \shortstack{\textbf{SCR}\\\textbf{Top-20}\\$\uparrow$} & \shortstack{\textbf{TPP}\\\textbf{Top-20}\\$\uparrow$} & \shortstack{\textbf{Sparse Probe}\\\textbf{Top-1}\\$\uparrow$} & \shortstack{\textbf{RAVEL}\\\textbf{Disent.}\\$\uparrow$} & \shortstack{\textbf{Sparsity}\\\textbf{(L0)}\\$\downarrow$} \\
\midrule
  \multirow{7}{*}{0} & \multirow{2}{*}{5M} & Random & 0.975 & \textbf{0.944} & 0.676 & 0.123 & 0.035 & 0.725 & 0.499 & 1208.2 \\
   &  & FPS & 0.955 & 0.942 & 0.680 & 0.105 & 0.021 & \textbf{0.735} & 0.499 & \textbf{1000.0} \\
  \cmidrule(lr){2-11}
   & \multirow{2}{*}{25M} & Random & \textbf{0.980} & 0.907 & 0.688 & 0.142 & 0.033 & 0.712 & 0.500 & 1200.3 \\
   &  & FPS & 0.960 & 0.942 & 0.689 & 0.120 & 0.029 & 0.727 & 0.501 & 1074.3 \\
  \cmidrule(lr){2-11}
   & \multirow{2}{*}{50M} & Random & 0.978 & 0.913 & 0.697 & \textbf{0.146} & 0.034 & 0.719 & 0.501 & 1180.7 \\
   &  & FPS & 0.962 & 0.918 & 0.702 & 0.127 & 0.030 & 0.724 & \textbf{0.502} & 1117.9 \\
  \cmidrule(lr){2-11}
   & \multirow{1}{*}{500M} & Baseline & 0.969 & 0.915 & \textbf{0.713} & 0.143 & \textbf{0.037} & 0.716 & 0.499 & 1144.5 \\
\bottomrule
\end{tabular}
}
\end{table*}

\begin{table*}[t]
\centering
\small
\setlength{\tabcolsep}{3.5pt}
\renewcommand{\arraystretch}{1.08}
\caption{\textbf{Per-trainer SAE evaluation for Pythia-160M, ReLU (64K).}}
\label{tab:pertrainer_pythia_standard_new_65536}
\resizebox{\textwidth}{!}{
\begin{tabular}{lllcccccccc}
\toprule
\textbf{Trainer} & \textbf{Budget} & \textbf{Selection} & \shortstack{\textbf{Loss}\\\textbf{Recovered}\\$\uparrow$} & \shortstack{\textbf{$1-$Abs.}\\\textbf{Fraction}\\$\uparrow$} & \shortstack{\textbf{AutoInterp}\\$\uparrow$} & \shortstack{\textbf{SCR}\\\textbf{Top-20}\\$\uparrow$} & \shortstack{\textbf{TPP}\\\textbf{Top-20}\\$\uparrow$} & \shortstack{\textbf{Sparse Probe}\\\textbf{Top-1}\\$\uparrow$} & \shortstack{\textbf{RAVEL}\\\textbf{Disent.}\\$\uparrow$} & \shortstack{\textbf{Sparsity}\\\textbf{(L0)}\\$\downarrow$} \\
\midrule
  \multirow{7}{*}{6} & \multirow{2}{*}{5M} & Random & 0.972 & \textbf{0.892} & 0.659 & 0.099 & 0.012 & 0.719 & 0.499 & 1861.4 \\
   &  & FPS & 0.944 & 0.878 & 0.666 & 0.076 & 0.008 & 0.719 & 0.495 & \textbf{1371.2} \\
  \cmidrule(lr){2-11}
   & \multirow{2}{*}{25M} & Random & 0.976 & 0.800 & 0.671 & 0.088 & 0.014 & 0.714 & 0.500 & 2002.8 \\
   &  & FPS & 0.964 & 0.784 & 0.679 & 0.090 & 0.010 & 0.706 & 0.499 & 1646.3 \\
  \cmidrule(lr){2-11}
   & \multirow{2}{*}{50M} & Random & \textbf{0.977} & 0.809 & 0.671 & 0.093 & 0.014 & 0.730 & 0.499 & 2012.6 \\
   &  & FPS & 0.967 & 0.728 & 0.674 & 0.083 & 0.012 & 0.718 & 0.498 & 1758.5 \\
  \cmidrule(lr){2-11}
   & \multirow{1}{*}{500M} & Baseline & 0.968 & 0.659 & \textbf{0.683} & \textbf{0.100} & \textbf{0.016} & \textbf{0.735} & \textbf{0.500} & 1997.5 \\
\bottomrule
\end{tabular}
}
\end{table*}

\begin{table*}[t]
\centering
\small
\setlength{\tabcolsep}{3.5pt}
\renewcommand{\arraystretch}{1.08}
\caption{\textbf{Per-trainer SAE evaluation for Pythia-160M, BatchTopK (16K).}}
\label{tab:pertrainer_pythia_batch_top_k_16384}
\resizebox{\textwidth}{!}{
\begin{tabular}{lllcccccccc}
\toprule
\textbf{Trainer} & \textbf{Budget} & \textbf{Selection} & \shortstack{\textbf{Loss}\\\textbf{Recovered}\\$\uparrow$} & \shortstack{\textbf{$1-$Abs.}\\\textbf{Fraction}\\$\uparrow$} & \shortstack{\textbf{AutoInterp}\\$\uparrow$} & \shortstack{\textbf{SCR}\\\textbf{Top-20}\\$\uparrow$} & \shortstack{\textbf{TPP}\\\textbf{Top-20}\\$\uparrow$} & \shortstack{\textbf{Sparse Probe}\\\textbf{Top-1}\\$\uparrow$} & \shortstack{\textbf{RAVEL}\\\textbf{Disent.}\\$\uparrow$} & \shortstack{\textbf{Sparsity}\\\textbf{(L0)}\\$\downarrow$} \\
\midrule
  \multirow{7}{*}{0} & \multirow{2}{*}{5M} & Random & 0.946 & 0.902 & 0.733 & 0.189 & \textbf{0.245} & 0.707 & 0.492 & 142.6 \\
   &  & FPS & 0.916 & 0.845 & 0.744 & \textbf{0.216} & 0.163 & \textbf{0.707} & 0.490 & \textbf{106.3} \\
  \cmidrule(lr){2-11}
   & \multirow{2}{*}{25M} & Random & 0.954 & 0.911 & 0.755 & 0.183 & 0.222 & 0.704 & 0.492 & 137.5 \\
   &  & FPS & 0.940 & 0.852 & 0.776 & 0.196 & 0.163 & 0.697 & 0.489 & 108.2 \\
  \cmidrule(lr){2-11}
   & \multirow{2}{*}{50M} & Random & 0.953 & \textbf{0.914} & 0.774 & 0.027 & 0.224 & 0.697 & 0.491 & 135.3 \\
   &  & FPS & 0.949 & 0.840 & 0.768 & 0.175 & 0.183 & 0.698 & 0.488 & 118.1 \\
  \cmidrule(lr){2-11}
   & \multirow{1}{*}{500M} & Baseline & \textbf{0.961} & 0.875 & \textbf{0.802} & -0.184 & 0.202 & 0.695 & \textbf{0.494} & 125.0 \\
\bottomrule
\end{tabular}
}
\end{table*}

\begin{table*}[t]
\centering
\small
\setlength{\tabcolsep}{3.5pt}
\renewcommand{\arraystretch}{1.08}
\caption{\textbf{Per-trainer SAE evaluation for Pythia-160M, BatchTopK (64K).}}
\label{tab:pertrainer_pythia_batch_top_k_65536}
\resizebox{\textwidth}{!}{
\begin{tabular}{lllcccccccc}
\toprule
\textbf{Trainer} & \textbf{Budget} & \textbf{Selection} & \shortstack{\textbf{Loss}\\\textbf{Recovered}\\$\uparrow$} & \shortstack{\textbf{$1-$Abs.}\\\textbf{Fraction}\\$\uparrow$} & \shortstack{\textbf{AutoInterp}\\$\uparrow$} & \shortstack{\textbf{SCR}\\\textbf{Top-20}\\$\uparrow$} & \shortstack{\textbf{TPP}\\\textbf{Top-20}\\$\uparrow$} & \shortstack{\textbf{Sparse Probe}\\\textbf{Top-1}\\$\uparrow$} & \shortstack{\textbf{RAVEL}\\\textbf{Disent.}\\$\uparrow$} & \shortstack{\textbf{Sparsity}\\\textbf{(L0)}\\$\downarrow$} \\
\midrule
  \multirow{7}{*}{2} & \multirow{2}{*}{5M} & Random & 0.945 & \textbf{0.886} & 0.692 & \textbf{0.229} & 0.248 & 0.708 & 0.492 & 352.4 \\
   &  & FPS & 0.916 & 0.836 & 0.703 & 0.189 & 0.210 & 0.704 & 0.488 & 294.7 \\
  \cmidrule(lr){2-11}
   & \multirow{2}{*}{25M} & Random & 0.961 & 0.862 & 0.717 & 0.200 & \textbf{0.260} & 0.697 & \textbf{0.492} & 330.6 \\
   &  & FPS & 0.946 & 0.769 & 0.763 & 0.196 & 0.215 & \textbf{0.710} & 0.483 & 276.4 \\
  \cmidrule(lr){2-11}
   & \multirow{2}{*}{50M} & Random & 0.961 & 0.857 & 0.736 & -0.084 & 0.235 & 0.709 & 0.490 & 315.7 \\
   &  & FPS & 0.952 & 0.635 & 0.746 & 0.180 & 0.145 & 0.691 & 0.488 & 282.9 \\
  \cmidrule(lr){2-11}
   & \multirow{1}{*}{500M} & Baseline & \textbf{0.961} & 0.782 & \textbf{0.783} & -0.061 & 0.224 & 0.684 & 0.490 & \textbf{267.7} \\
\bottomrule
\end{tabular}
}
\end{table*}

\begin{table*}[t]
\centering
\small
\setlength{\tabcolsep}{3.5pt}
\renewcommand{\arraystretch}{1.08}
\caption{\textbf{Per-trainer SAE evaluation for Pythia-160M, JumpReLU (16K).}}
\label{tab:pertrainer_pythia_jump_relu_16384}
\resizebox{\textwidth}{!}{
\begin{tabular}{lllcccccccc}
\toprule
\textbf{Trainer} & \textbf{Budget} & \textbf{Selection} & \shortstack{\textbf{Loss}\\\textbf{Recovered}\\$\uparrow$} & \shortstack{\textbf{$1-$Abs.}\\\textbf{Fraction}\\$\uparrow$} & \shortstack{\textbf{AutoInterp}\\$\uparrow$} & \shortstack{\textbf{SCR}\\\textbf{Top-20}\\$\uparrow$} & \shortstack{\textbf{TPP}\\\textbf{Top-20}\\$\uparrow$} & \shortstack{\textbf{Sparse Probe}\\\textbf{Top-1}\\$\uparrow$} & \shortstack{\textbf{RAVEL}\\\textbf{Disent.}\\$\uparrow$} & \shortstack{\textbf{Sparsity}\\\textbf{(L0)}\\$\downarrow$} \\
\midrule
  \multirow{7}{*}{0} & \multirow{2}{*}{5M} & Random & 0.946 & 0.857 & 0.732 & 0.240 & 0.229 & 0.696 & 0.491 & 139.5 \\
   &  & FPS & 0.909 & 0.809 & 0.747 & 0.131 & 0.210 & 0.691 & 0.491 & \textbf{106.7} \\
  \cmidrule(lr){2-11}
   & \multirow{2}{*}{25M} & Random & 0.955 & 0.907 & 0.742 & \textbf{0.246} & 0.213 & 0.686 & 0.486 & 141.6 \\
   &  & FPS & 0.942 & 0.832 & 0.763 & 0.161 & 0.175 & 0.686 & 0.490 & 111.9 \\
  \cmidrule(lr){2-11}
   & \multirow{2}{*}{50M} & Random & \textbf{0.957} & \textbf{0.908} & 0.753 & 0.208 & \textbf{0.246} & 0.695 & 0.492 & 141.5 \\
   &  & FPS & 0.948 & 0.828 & \textbf{0.766} & 0.214 & 0.186 & 0.680 & 0.488 & 115.6 \\
  \cmidrule(lr){2-11}
   & \multirow{1}{*}{500M} & Baseline & 0.956 & 0.889 & 0.762 & 0.204 & 0.199 & \textbf{0.711} & \textbf{0.493} & 138.0 \\
\bottomrule
\end{tabular}
}
\end{table*}

\begin{table*}[t]
\centering
\small
\setlength{\tabcolsep}{3.5pt}
\renewcommand{\arraystretch}{1.08}
\caption{\textbf{Per-trainer SAE evaluation for Pythia-160M, JumpReLU (64K).}}
\label{tab:pertrainer_pythia_jump_relu_65536}
\resizebox{\textwidth}{!}{
\begin{tabular}{lllcccccccc}
\toprule
\textbf{Trainer} & \textbf{Budget} & \textbf{Selection} & \shortstack{\textbf{Loss}\\\textbf{Recovered}\\$\uparrow$} & \shortstack{\textbf{$1-$Abs.}\\\textbf{Fraction}\\$\uparrow$} & \shortstack{\textbf{AutoInterp}\\$\uparrow$} & \shortstack{\textbf{SCR}\\\textbf{Top-20}\\$\uparrow$} & \shortstack{\textbf{TPP}\\\textbf{Top-20}\\$\uparrow$} & \shortstack{\textbf{Sparse Probe}\\\textbf{Top-1}\\$\uparrow$} & \shortstack{\textbf{RAVEL}\\\textbf{Disent.}\\$\uparrow$} & \shortstack{\textbf{Sparsity}\\\textbf{(L0)}\\$\downarrow$} \\
\midrule
  \multirow{7}{*}{2} & \multirow{2}{*}{5M} & Random & 0.950 & \textbf{0.833} & 0.703 & 0.147 & 0.211 & 0.698 & 0.494 & 297.4 \\
   &  & FPS & 0.909 & 0.739 & 0.714 & 0.125 & 0.165 & 0.665 & 0.490 & \textbf{255.9} \\
  \cmidrule(lr){2-11}
   & \multirow{2}{*}{25M} & Random & 0.958 & 0.818 & 0.727 & \textbf{0.244} & \textbf{0.238} & 0.698 & 0.494 & 309.8 \\
   &  & FPS & 0.944 & 0.654 & 0.733 & 0.163 & 0.168 & \textbf{0.722} & 0.487 & 266.2 \\
  \cmidrule(lr){2-11}
   & \multirow{2}{*}{50M} & Random & 0.961 & 0.780 & 0.725 & 0.218 & 0.205 & 0.676 & 0.492 & 315.0 \\
   &  & FPS & 0.957 & 0.635 & 0.747 & 0.207 & 0.184 & 0.690 & \textbf{0.494} & 272.5 \\
  \cmidrule(lr){2-11}
   & \multirow{1}{*}{500M} & Baseline & \textbf{0.963} & 0.737 & \textbf{0.752} & 0.162 & 0.191 & 0.700 & 0.487 & 322.2 \\
\bottomrule
\end{tabular}
}
\end{table*}

\begin{table*}[t]
\centering
\small
\setlength{\tabcolsep}{3.5pt}
\renewcommand{\arraystretch}{1.08}
\caption{\textbf{Per-trainer SAE evaluation for Pythia-160M, Matryoshka (16K).}}
\label{tab:pertrainer_pythia_matryoshka_batch_top_k_16384}
\resizebox{\textwidth}{!}{
\begin{tabular}{lllcccccccc}
\toprule
\textbf{Trainer} & \textbf{Budget} & \textbf{Selection} & \shortstack{\textbf{Loss}\\\textbf{Recovered}\\$\uparrow$} & \shortstack{\textbf{$1-$Abs.}\\\textbf{Fraction}\\$\uparrow$} & \shortstack{\textbf{AutoInterp}\\$\uparrow$} & \shortstack{\textbf{SCR}\\\textbf{Top-20}\\$\uparrow$} & \shortstack{\textbf{TPP}\\\textbf{Top-20}\\$\uparrow$} & \shortstack{\textbf{Sparse Probe}\\\textbf{Top-1}\\$\uparrow$} & \shortstack{\textbf{RAVEL}\\\textbf{Disent.}\\$\uparrow$} & \shortstack{\textbf{Sparsity}\\\textbf{(L0)}\\$\downarrow$} \\
\midrule
  \multirow{7}{*}{0} & \multirow{2}{*}{5M} & Random & 0.934 & 0.902 & 0.758 & 0.136 & 0.244 & \textbf{0.727} & 0.486 & 132.5 \\
   &  & FPS & 0.914 & 0.892 & 0.755 & \textbf{0.182} & 0.223 & 0.693 & 0.490 & 107.5 \\
  \cmidrule(lr){2-11}
   & \multirow{2}{*}{25M} & Random & \textbf{0.952} & 0.907 & 0.791 & -0.040 & \textbf{0.277} & 0.720 & \textbf{0.493} & 126.6 \\
   &  & FPS & 0.931 & 0.938 & 0.795 & -0.086 & 0.238 & 0.699 & 0.493 & \textbf{101.8} \\
  \cmidrule(lr){2-11}
   & \multirow{2}{*}{50M} & Random & 0.949 & 0.890 & 0.803 & -0.302 & 0.213 & 0.697 & 0.489 & 120.6 \\
   &  & FPS & 0.941 & \textbf{0.952} & 0.800 & 0.037 & 0.232 & 0.714 & 0.490 & 107.5 \\
  \cmidrule(lr){2-11}
   & \multirow{1}{*}{500M} & Baseline & 0.951 & 0.942 & \textbf{0.819} & -0.187 & 0.232 & 0.702 & 0.487 & 117.6 \\
\bottomrule
\end{tabular}
}
\end{table*}

\begin{table*}[t]
\centering
\small
\setlength{\tabcolsep}{3.5pt}
\renewcommand{\arraystretch}{1.08}
\caption{\textbf{Per-trainer SAE evaluation for Pythia-160M, Matryoshka (64K).}}
\label{tab:pertrainer_pythia_matryoshka_batch_top_k_65536}
\resizebox{\textwidth}{!}{
\begin{tabular}{lllcccccccc}
\toprule
\textbf{Trainer} & \textbf{Budget} & \textbf{Selection} & \shortstack{\textbf{Loss}\\\textbf{Recovered}\\$\uparrow$} & \shortstack{\textbf{$1-$Abs.}\\\textbf{Fraction}\\$\uparrow$} & \shortstack{\textbf{AutoInterp}\\$\uparrow$} & \shortstack{\textbf{SCR}\\\textbf{Top-20}\\$\uparrow$} & \shortstack{\textbf{TPP}\\\textbf{Top-20}\\$\uparrow$} & \shortstack{\textbf{Sparse Probe}\\\textbf{Top-1}\\$\uparrow$} & \shortstack{\textbf{RAVEL}\\\textbf{Disent.}\\$\uparrow$} & \shortstack{\textbf{Sparsity}\\\textbf{(L0)}\\$\downarrow$} \\
\midrule
  \multirow{7}{*}{2} & \multirow{2}{*}{5M} & Random & 0.948 & 0.891 & 0.687 & 0.120 & 0.214 & \textbf{0.717} & 0.490 & 340.3 \\
   &  & FPS & 0.923 & 0.877 & 0.690 & \textbf{0.209} & 0.214 & 0.717 & 0.490 & 314.7 \\
  \cmidrule(lr){2-11}
   & \multirow{2}{*}{25M} & Random & 0.954 & 0.939 & 0.712 & -0.259 & \textbf{0.251} & 0.714 & 0.492 & 330.7 \\
   &  & FPS & 0.945 & 0.941 & 0.734 & -0.035 & 0.192 & 0.704 & 0.489 & 283.3 \\
  \cmidrule(lr){2-11}
   & \multirow{2}{*}{50M} & Random & 0.954 & \textbf{0.967} & 0.744 & -0.085 & 0.219 & 0.694 & 0.489 & 300.2 \\
   &  & FPS & 0.951 & 0.942 & 0.747 & 0.011 & 0.173 & 0.710 & 0.489 & \textbf{270.7} \\
  \cmidrule(lr){2-11}
   & \multirow{1}{*}{500M} & Baseline & \textbf{0.958} & 0.950 & \textbf{0.763} & -0.154 & 0.228 & 0.706 & \textbf{0.493} & 279.1 \\
\bottomrule
\end{tabular}
}
\end{table*}


\begin{table*}[t]
\centering
\small
\setlength{\tabcolsep}{3.5pt}
\renewcommand{\arraystretch}{1.08}
\caption{\textbf{Per-trainer SAE evaluation for Qwen3-8B, ReLU (64K).}}
\label{tab:pertrainer_qwen3_standard_new_65536}
\resizebox{\textwidth}{!}{
\begin{tabular}{lllccccccc}
\toprule
\textbf{Trainer} & \textbf{Budget} & \textbf{Selection} & \shortstack{\textbf{Loss}\\\textbf{Recovered}\\$\uparrow$} & \shortstack{\textbf{$1-$Abs.}\\\textbf{Fraction}\\$\uparrow$} & \shortstack{\textbf{AutoInterp}\\$\uparrow$} & \shortstack{\textbf{SCR}\\\textbf{Top-20}\\$\uparrow$} & \shortstack{\textbf{TPP}\\\textbf{Top-20}\\$\uparrow$} & \shortstack{\textbf{Sparse Probe}\\\textbf{Top-1}\\$\uparrow$} & \shortstack{\textbf{Sparsity}\\\textbf{(L0)}\\$\downarrow$} \\
\midrule
  \multirow{7}{*}{0} & \multirow{2}{*}{5M} & Random & 0.993 & 0.959 & 0.823 & \textbf{0.178} & \textbf{0.155} & 0.732 & 1337.0 \\
   &  & FPS & 0.984 & \textbf{0.970} & 0.808 & 0.116 & 0.057 & 0.732 & \textbf{1199.0} \\
  \cmidrule(lr){2-10}
   & \multirow{2}{*}{25M} & Random & 0.993 & 0.894 & \textbf{0.833} & 0.147 & 0.101 & 0.726 & 1835.8 \\
   &  & FPS & 0.991 & 0.860 & 0.812 & 0.137 & 0.065 & 0.732 & 1638.7 \\
  \cmidrule(lr){2-10}
   & \multirow{2}{*}{50M} & Random & \textbf{0.995} & 0.815 & 0.815 & 0.151 & 0.090 & \textbf{0.780} & 1944.7 \\
   &  & FPS & 0.991 & 0.755 & 0.793 & 0.125 & 0.054 & 0.737 & 1703.2 \\
  \cmidrule(lr){2-10}
   & \multirow{1}{*}{500M} & Baseline & \textbf{0.995} & 0.734 & 0.777 & 0.146 & 0.066 & 0.666 & 2003.9 \\
\bottomrule
\end{tabular}
}
\end{table*}

\begin{table*}[t]
\centering
\small
\setlength{\tabcolsep}{3.5pt}
\renewcommand{\arraystretch}{1.08}
\caption{\textbf{Per-trainer SAE evaluation for Qwen3-8B, BatchTopK (64K).}}
\label{tab:pertrainer_qwen3_batch_top_k_65536}
\resizebox{\textwidth}{!}{
\begin{tabular}{lllccccccc}
\toprule
\textbf{Trainer} & \textbf{Budget} & \textbf{Selection} & \shortstack{\textbf{Loss}\\\textbf{Recovered}\\$\uparrow$} & \shortstack{\textbf{$1-$Abs.}\\\textbf{Fraction}\\$\uparrow$} & \shortstack{\textbf{AutoInterp}\\$\uparrow$} & \shortstack{\textbf{SCR}\\\textbf{Top-20}\\$\uparrow$} & \shortstack{\textbf{TPP}\\\textbf{Top-20}\\$\uparrow$} & \shortstack{\textbf{Sparse Probe}\\\textbf{Top-1}\\$\uparrow$} & \shortstack{\textbf{Sparsity}\\\textbf{(L0)}\\$\downarrow$} \\
\midrule
  \multirow{7}{*}{2} & \multirow{2}{*}{5M} & Random & 0.984 & 0.974 & 0.852 & \textbf{0.282} & 0.303 & \textbf{0.703} & 85.6 \\
   &  & FPS & 0.968 & 0.962 & 0.878 & 0.189 & 0.258 & 0.679 & \textbf{63.4} \\
  \cmidrule(lr){2-10}
   & \multirow{2}{*}{25M} & Random & 0.986 & 0.951 & 0.867 & 0.264 & 0.303 & 0.692 & 86.1 \\
   &  & FPS & 0.977 & \textbf{0.983} & \textbf{0.882} & 0.202 & 0.296 & 0.671 & 66.9 \\
  \cmidrule(lr){2-10}
   & \multirow{2}{*}{50M} & Random & 0.986 & 0.944 & 0.867 & 0.258 & 0.289 & 0.692 & 85.8 \\
   &  & FPS & 0.982 & 0.943 & 0.877 & 0.236 & \textbf{0.313} & 0.698 & 69.6 \\
  \cmidrule(lr){2-10}
   & \multirow{1}{*}{500M} & Baseline & \textbf{0.989} & 0.880 & 0.866 & 0.180 & 0.247 & 0.669 & 87.2 \\
\bottomrule
\end{tabular}
}
\end{table*}

\begin{table*}[t]
\centering
\small
\setlength{\tabcolsep}{3.5pt}
\renewcommand{\arraystretch}{1.08}
\caption{\textbf{Per-trainer SAE evaluation for Qwen3-8B, JumpReLU (64K).}}
\label{tab:pertrainer_qwen3_jump_relu_65536}
\resizebox{\textwidth}{!}{
\begin{tabular}{lllccccccc}
\toprule
\textbf{Trainer} & \textbf{Budget} & \textbf{Selection} & \shortstack{\textbf{Loss}\\\textbf{Recovered}\\$\uparrow$} & \shortstack{\textbf{$1-$Abs.}\\\textbf{Fraction}\\$\uparrow$} & \shortstack{\textbf{AutoInterp}\\$\uparrow$} & \shortstack{\textbf{SCR}\\\textbf{Top-20}\\$\uparrow$} & \shortstack{\textbf{TPP}\\\textbf{Top-20}\\$\uparrow$} & \shortstack{\textbf{Sparse Probe}\\\textbf{Top-1}\\$\uparrow$} & \shortstack{\textbf{Sparsity}\\\textbf{(L0)}\\$\downarrow$} \\
\midrule
  \multirow{7}{*}{2} & \multirow{2}{*}{5M} & Random & 0.980 & \textbf{0.990} & 0.857 & \textbf{0.336} & 0.226 & 0.652 & 87.4 \\
   &  & FPS & 0.968 & 0.943 & 0.860 & 0.192 & 0.106 & \textbf{0.700} & 68.5 \\
  \cmidrule(lr){2-10}
   & \multirow{2}{*}{25M} & Random & 0.984 & 0.932 & 0.869 & 0.270 & \textbf{0.229} & 0.664 & 87.3 \\
   &  & FPS & 0.982 & 0.965 & \textbf{0.884} & 0.212 & 0.098 & 0.677 & \textbf{67.7} \\
  \cmidrule(lr){2-10}
   & \multirow{2}{*}{50M} & Random & 0.986 & 0.962 & 0.876 & 0.257 & 0.198 & 0.671 & 87.4 \\
   &  & FPS & 0.982 & 0.909 & 0.868 & 0.242 & 0.151 & 0.662 & 70.1 \\
  \cmidrule(lr){2-10}
   & \multirow{1}{*}{500M} & Baseline & \textbf{0.987} & 0.833 & 0.867 & 0.206 & 0.132 & 0.672 & 88.3 \\
\bottomrule
\end{tabular}
}
\end{table*}

\begin{table*}[t]
\centering
\small
\setlength{\tabcolsep}{3.5pt}
\renewcommand{\arraystretch}{1.08}
\caption{\textbf{Per-trainer SAE evaluation for Qwen3-8B, Matryoshka (64K).}}
\label{tab:pertrainer_qwen3_matryoshka_batch_top_k_65536}
\resizebox{\textwidth}{!}{
\begin{tabular}{lllccccccc}
\toprule
\textbf{Trainer} & \textbf{Budget} & \textbf{Selection} & \shortstack{\textbf{Loss}\\\textbf{Recovered}\\$\uparrow$} & \shortstack{\textbf{$1-$Abs.}\\\textbf{Fraction}\\$\uparrow$} & \shortstack{\textbf{AutoInterp}\\$\uparrow$} & \shortstack{\textbf{SCR}\\\textbf{Top-20}\\$\uparrow$} & \shortstack{\textbf{TPP}\\\textbf{Top-20}\\$\uparrow$} & \shortstack{\textbf{Sparse Probe}\\\textbf{Top-1}\\$\uparrow$} & \shortstack{\textbf{Sparsity}\\\textbf{(L0)}\\$\downarrow$} \\
\midrule
  \multirow{7}{*}{2} & \multirow{2}{*}{5M} & Random & 0.980 & 0.982 & 0.848 & 0.290 & \textbf{0.342} & 0.662 & 85.3 \\
   &  & FPS & 0.962 & \textbf{0.998} & 0.892 & 0.238 & 0.286 & \textbf{0.726} & \textbf{58.3} \\
  \cmidrule(lr){2-10}
   & \multirow{2}{*}{25M} & Random & 0.984 & 0.990 & 0.863 & 0.268 & 0.325 & 0.708 & 85.5 \\
   &  & FPS & 0.978 & 0.986 & \textbf{0.894} & 0.296 & 0.298 & 0.672 & 62.6 \\
  \cmidrule(lr){2-10}
   & \multirow{2}{*}{50M} & Random & \textbf{0.986} & 0.986 & 0.857 & 0.365 & 0.311 & 0.691 & 85.9 \\
   &  & FPS & 0.980 & 0.978 & 0.894 & 0.312 & 0.299 & 0.704 & 65.7 \\
  \cmidrule(lr){2-10}
   & \multirow{1}{*}{500M} & Baseline & 0.984 & 0.974 & 0.850 & \textbf{0.372} & 0.314 & 0.705 & 86.1 \\
\bottomrule
\end{tabular}
}
\end{table*}
\clearpage
\section{Proofs: When Should Feature matchings Agree?}
\label{appendix:feature_alignment_theory}

\subsection{Separate encoder and decoder feature alignment~\ref{thm:encoder_decoder_alignment}}

\begin{proof}
Define
\[
a_{r,N}
=
\sup_\theta
\left|
\widehat R_{r,N}(\theta)-R(\theta)
\right|.
\]
Statistical consistency (assumption (ii)) implies
$a_{1,N}\xrightarrow{p}0$ and $a_{q,N}\xrightarrow{p}0$.

Let $\theta^\star$ be a population minimizer. Approximate empirical
optimality gives
\begin{align}
R(\widehat\theta_{r,N})
&\leq
\widehat R_{r,N}(\widehat\theta_{r,N})
+
a_{r,N}
\\
&\leq
\widehat R_{r,N}(\theta^\star)
+
\eta_{r,N}
+
a_{r,N}
\\
&\leq
R^\star
+
2a_{r,N}
+
\eta_{r,N}.
\end{align}
Here $\eta_{r,N}\ge 0$ denotes the optimization error, i.e.,
$\widehat R_{r,N}(\widehat\theta_{r,N})\le\inf_\theta\widehat R_{r,N}(\theta)+\eta_{r,N}$,
and $\eta_{r,N}\xrightarrow{p}0$ by optimization consistency (assumption (iv)).


Hence
\[
R(\widehat\theta_{r,N})-R^\star
\xrightarrow{p}0.
\]

By encoder identifiability and separation, any sequence whose encoder
features remain at least $\epsilon$ away from the population encoder orbit
must incur population excess risk at least
$\Delta_E(\epsilon)>0$. Therefore
\[
\Pr\left[
d_E(\widehat\theta_{r,N},\theta^\star)
\geq\epsilon
\right]
\leq
\Pr\left[
2a_{r,N}+\eta_{r,N}
\geq
\Delta_E(\epsilon)
\right]
\rightarrow0.
\]

The same argument using decoder separation gives
$
d_D(\widehat\theta_{r,N},\theta^\star)\xrightarrow{p}0.
$
Applying the triangle inequality separately in encoder and decoder feature
spaces gives
\[
d_E(
\widehat\theta_{q,N},
\widehat\theta_{1,N}
)
\xrightarrow{p}0
\]
and
\[
d_D(
\widehat\theta_{q,N},
\widehat\theta_{1,N}
)
\xrightarrow{p}0.
\]
\end{proof}

\subsection{Proof of Agreement of encoder and decoder Hungarian matchings) ~\ref{cor:assignment_vs_similarity}}

\begin{proof}
By Theorem~\ref{thm:encoder_decoder_alignment}, both normalized encoder and
decoder feature systems converge to their respective population feature
systems after one-to-one alignment.



By assumption (iii), each run's encoder and decoder features approach the
population features under a single labeling shared by encoder and decoder.
Let $P^\star_N$ denote the permutation that aligns the two runs' labelings.
Because distinct population features are separated by a positive margin,
small perturbations cannot change the optimal Hungarian assignment, so with
probability tending to one $P_E^{(N)}=P^\star_N$, and likewise $P_D^{(N)}=P^\star_N$.

Hence,
\[
\Pr
\left(
P_E^{(N)}
=
P_D^{(N)}
\right)
\rightarrow1.
\]
\end{proof}

\begin{table*}[t]
\centering
\small
\setlength{\tabcolsep}{2.5pt}
\renewcommand{\arraystretch}{0.9}
\caption{
\textbf{FPS distribution shift for Qwen3-8B and Pythia-160M.\textsuperscript{*}}
Metrics are defined as in Table~\ref{tab:fps_distribution_shift}. The Gemma-2-2B pattern holds across model families: random selection is distribution-neutral at every budget, while FPS shifts the selected subset toward higher-norm activations, with divergence growing as the subset shrinks.
}
\label{tab:fps_distribution_shift_appendix}
\begin{tabular}{lccccc}
\toprule
&&&
\multicolumn{3}{c}{\textbf{Selected vs.\ 500M stream}} \\
\cmidrule(lr){4-6}

\textbf{Model}
& \textbf{Selection}
& \textbf{Coreset}
& \multicolumn{1}{c}{$\boldsymbol{\Delta}$\textbf{ mean norm (\%)}}
& \multicolumn{1}{c}{\textbf{KL (bits)}}
& \multicolumn{1}{c}{\textbf{Mean-dir.\ cosine}} \\
\midrule

\multirow{6}{*}{Qwen3-8B}
& \multirow{3}{*}{FPS}
& 5M  & $+33.8 \pm 1.3$ & 4.02 & $0.889 \pm 0.007$ \\
&& 25M & $+21.4 \pm 0.4$ & 2.08 & $0.952 \pm 0.002$ \\
&& 50M & $+16.4 \pm 0.3$ & 1.35 & $0.970 \pm 0.001$ \\
\cmidrule(lr){2-6}
& \multirow{3}{*}{Random}
& 5M  & $+0.1 \pm 0.2$ & 0.000 & $1.000 \pm 0.000$ \\
&& 25M & $0.0 \pm 0.1$  & 0.000 & $1.000 \pm 0.000$ \\
&& 50M & $0.0 \pm 0.1$  & 0.000 & $1.000 \pm 0.000$ \\

\midrule

\multirow{6}{*}{Pythia-160M}
& \multirow{3}{*}{FPS}
& 5M  & $+23.7 \pm 0.5$ & 1.49 & $0.900 \pm 0.004$ \\
&& 25M & $+15.5 \pm 0.3$ & 0.78 & $0.953 \pm 0.002$ \\
&& 50M & $+11.9 \pm 0.3$ & 0.51 & $0.973 \pm 0.001$ \\
\cmidrule(lr){2-6}
& \multirow{3}{*}{Random}
& 5M  & $0.0 \pm 0.3$  & 0.000 & $1.000 \pm 0.000$ \\
&& 25M & $0.0 \pm 0.1$  & 0.000 & $1.000 \pm 0.000$ \\
&& 50M & $0.0 \pm 0.1$  & 0.000 & $1.000 \pm 0.000$ \\

\bottomrule
\end{tabular}

\vspace{2pt}
{\scriptsize *Same protocol as Table~\ref{tab:fps_distribution_shift}: all selectors are applied to identical activation pools; entries are mean $\pm$ SD across pool refreshes (50 for norm statistics, 15 for direction statistics). Entries shown as $\pm 0.000$ have SD below $10^{-4}$.\par}
\end{table*}

\begin{table*}[t]
\centering
\small
\setlength{\tabcolsep}{2.5pt}
\renewcommand{\arraystretch}{0.9}
\caption{\textbf{Feature correspondence under increasingly strict alignment
criteria.}
\emph{Encoder-decoder agreement} is the fraction of features for which
independently computed encoder and decoder Hungarian matchings select the same
reference latent. \emph{Shared features} additionally require both encoder and
decoder cosine similarities to exceed the specified threshold $\tau$. Results
are reported for Qwen3-8B, and Pythia-160M.}
\label{tab:feature_correspondence_all}

\begin{tabular}{llccccc}
\toprule
&&&
\multicolumn{1}{c}{\textbf{Counterpart (\%)}}
&
\multicolumn{3}{c}{\textbf{Shared features (\%)}} \\
\cmidrule(lr){4-4}
\cmidrule(lr){5-7}
\textbf{Model}
& \textbf{SAE}
& \textbf{Budget}
& \textbf{Enc.--dec. agree}
& $\boldsymbol{\tau \ge 0.3}$
& $\boldsymbol{\tau \ge 0.5}$
& $\boldsymbol{\tau \ge 0.7}$ \\
\midrule



\multirow{6}{*}{Qwen3-8B}
& \multirow{3}{*}{ReLU}
& 5M  & 69.85 & 33.07 & 22.90 & 18.58 \\
&& 25M & 79.47 & 51.16 & 36.23 & 25.19 \\
&& 50M & \textbf{82.49} & \textbf{61.70} & \textbf{41.41} & \textbf{29.51}\\
\cmidrule(lr){2-7}
& \multirow{3}{*}{BatchTopK}
& 5M  & 66.06 & 34.16 &  7.34 &  1.14 \\
&& 25M & 81.05 & 65.56 & 35.63 &  8.83 \\
&& 50M & \textbf{84.69} & \textbf{72.78} & \textbf{49.49} & \textbf{18.24} \\

\midrule

\multirow{6}{*}{Pythia-160M}
& \multirow{3}{*}{ReLU}
& 5M  & 73.90 & 47.81 & 29.79 & 23.54 \\
&& 25M & 78.11 & 61.07 & 37.96 & 29.09 \\
&& 50M & \textbf{79.85} & \textbf{65.89}
        & \textbf{42.46} & \textbf{32.16} \\
\cmidrule(lr){2-7}
& \multirow{3}{*}{BatchTopK}
& 5M  & 47.19 & 31.81 &  7.80 &  1.72 \\
&& 25M & 64.92 & 55.84 & 28.81 &  8.37 \\
&& 50M & \textbf{69.35} & \textbf{61.61}
        & \textbf{38.13} & \textbf{14.86} \\

\bottomrule
\end{tabular}
\end{table*}

\begin{table*}[t]
\centering
\scriptsize
\setlength{\tabcolsep}{2.2pt}
\renewcommand{\arraystretch}{1.0}

\caption{
\textbf{Paired statistical tests for Gemma-2-2B ReLU and BatchTopK.}
Panel A compares the \(500\)M reference with Random subsets
(\(\Delta=\mathrm{500M}-\mathrm{Random}\)); Panel B compares FPS with Random
at matched budgets (\(\Delta=\mathrm{FPS}-\mathrm{Random}\)).
Each entry reports \(\Delta/p\) from a paired two-sided \(t\)-test across
matched SAE configurations.
}
\label{tab:stats_gemma}

\begin{tabular}{lccc@{\hspace{4pt}}ccc}
\toprule
&
\multicolumn{3}{c}{\textbf{ReLU}}
&
\multicolumn{3}{c}{\textbf{BatchTopK}} \\
\cmidrule(lr){2-4}
\cmidrule(lr){5-7}

\textbf{Metric}
& \textbf{5M} & \textbf{25M} & \textbf{50M}
& \textbf{5M} & \textbf{25M} & \textbf{50M} \\
\midrule

\multicolumn{7}{l}{\textbf{A. 500M vs.\ Random}} \\
\midrule

Loss Rec.
& \(+.005/<.001^{***}\)
& \(+.002/.002^{**}\)
& \(+.002/.014^{*}\)
& \(+.007/.006^{**}\)
& \(+.002/.015^{*}\)
& \(+.002/.092\) \\

Sparse Probe
& \(+.018/.008^{**}\)
& \(+.016/.005^{**}\)
& \(+.004/.363\)
& \(-.022/.355\)
& \(-.036/.050\)
& \(-.006/.769\) \\

1-Abs.
& \(-.098/<.001^{***}\)
& \(-.053/<.001^{***}\)
& \(-.030/<.001^{***}\)
& \(-.032/.516\)
& \(-.047/.252\)
& \(-.055/.135\) \\

AutoInterp
& \(+.020/<.001^{***}\)
& \(+.014/<.001^{***}\)
& \(+.006/.085\)
& \(+.032/.039^{*}\)
& \(+.008/.365\)
& \(+.003/.734\) \\

SCR
& \(+.007/.287\)
& \(+.002/.765\)
& \(+.003/.454\)
& \(-.055/.173\)
& \(-.037/.239\)
& \(-.034/.133\) \\

TPP
& \(+.001/.016^{*}\)
& \(+.001/.031^{*}\)
& \(+.000/.616\)
& \(-.023/.080\)
& \(-.040/.136\)
& \(-.018/.225\) \\

RAVEL
& \(+.016/.123\)
& \(+.006/.260\)
& \(+.005/.555\)
& \(+.024/.006^{**}\)
& \(+.007/.539\)
& \(+.009/.231\) \\

Unlearning
& \(+.009/.602\)
& \(+.002/.902\)
& \(-.008/.418\)
& \(+.060/.076\)
& \(+.044/.243\)
& \(+.047/.005^{**}\) \\

\midrule
\multicolumn{7}{l}{\textbf{B. FPS vs.\ Random}} \\
\midrule

Loss Rec.
& \(-.020/<.001^{***}\)
& \(-.010/<.001^{***}\)
& \(-.007/.001^{**}\)
& \(-.023/.002^{**}\)
& \(-.009/.009^{**}\)
& \(-.005/.010^{*}\) \\

Sparse Probe
& \(-.016/.027^{*}\)
& \(-.005/.425\)
& \(-.019/.007^{**}\)
& \(-.013/.108\)
& \(-.026/.346\)
& \(-.017/.346\) \\

1-Abs.
& \(-.011/.242\)
& \(-.063/.001^{**}\)
& \(-.054/<.001^{***}\)
& \(-.010/.169\)
& \(-.108/.104\)
& \(-.126/.049^{*}\) \\

AutoInterp
& \(+.006/.014^{*}\)
& \(+.014/<.001^{***}\)
& \(+.010/.023^{*}\)
& \(+.026/.143\)
& \(+.027/.019^{*}\)
& \(+.025/.007^{**}\) \\

SCR
& \(-.064/<.001^{***}\)
& \(-.050/<.001^{***}\)
& \(-.037/<.001^{***}\)
& \(-.090/.004^{**}\)
& \(-.067/.041^{*}\)
& \(-.084/.084\) \\

TPP
& \(-.006/<.001^{***}\)
& \(-.004/<.001^{***}\)
& \(-.004/<.001^{***}\)
& \(-.022/.006^{**}\)
& \(-.032/.070\)
& \(-.001/.982\) \\

RAVEL
& \(-.012/.053\)
& \(-.011/.016^{*}\)
& \(-.010/.215\)
& \(-.029/.006^{**}\)
& \(-.015/.138\)
& \(-.003/.642\) \\

Unlearning
& \(+.016/.366\)
& \(+.006/.673\)
& \(+.015/.380\)
& \(+.022/.123\)
& \(+.019/.593\)
& \(+.014/.351\) \\

\bottomrule
\end{tabular}

\vspace{2pt}
{\scriptsize
\textit{Observation.}
Reduced-data SAEs become increasingly comparable to the \(500\)M reference as
the budget grows, with several metrics already showing parity at small budgets.
Selection effects are metric-dependent: Random generally favors reconstruction
and SCR, whereas FPS frequently improves AutoInterp.
\(^{***}p<0.001\), \(^{**}p<0.01\), \(^{*}p<0.05\).
}

\end{table*}

\end{document}